\documentclass{article} 
\usepackage{iclr2023_conference,times}

\usepackage[toc,page,header]{appendix}
\usepackage{tocbasic}[2020/07/22]

\def\ceil#1{\lceil #1 \rceil}
\def\floor#1{\lfloor #1 \rfloor}
\def\1{\bm{1}}

\usepackage{floatrow}
\newfloatcommand{capbtabbox}{table}[][\FBwidth]

\usepackage{blindtext}

\usepackage[colorlinks]{hyperref}       
\hypersetup{colorlinks,allcolors=[rgb]{0,0.07843,0.45098}}

\usepackage{comment}
\usepackage{titlesec}
\titlespacing{\section}{0pt}{1pt}{-1pt}
\titlespacing{\subsection}{0pt}{1pt}{-1pt}
\titlespacing{\subsubsection}{0pt}{1pt}{-1pt}

\usepackage{url}
\usepackage{kz_style}   
\newcommand{\kz}[1]{\textcolor{olive}{[Kaiqing: #1]}}

\newcommand{\lirui}[1]{\textcolor{blue}{[Lirui: #1]}}

\usepackage{floatrow}

\newcommand{\apbbox}[1]{AP$^\text{bb}_\text{#1}$}
\newcommand{\apmask}[1]{AP$^\text{mk}_\text{#1}$}
\definecolor{Gray}{gray}{0.5}
\definecolor{nicergreen}{rgb}{0.13, 0.54, 0.13}
\definecolor{nicered}{rgb}{0.83, 0.16, 0.16}
\definecolor{Highlight}{HTML}{39b54a}  

\newcommand{\cgaphl}[2]{
\fontsize{6pt}{1em}\selectfont{\textcolor{nicergreen}{(${#1}$\textbf{#2})}} 
}

\title{
 {Does learning from decentralized non-IID  \\
unlabeled 
data benefit from  self supervision?}}

\author{Lirui Wang,
Kaiqing Zhang,
Yunzhu Li,
Yonglong Tian,
Russ Tedrake\\
MIT CSAIL\\
} 

  \iclrfinalcopy 
\begin{document}



\maketitle

\begin{abstract}
 The success of machine learning relies heavily on massive amounts of data, which are usually generated and stored across a range of diverse and distributed data sources. {\it Decentralized learning} has thus been advocated and widely deployed to make efficient use of the distributed datasets, with an extensive focus on supervised learning (SL) problems.  Unfortunately, the majority of real-world data are {\it unlabeled} and can be highly {\it heterogeneous} across sources. In this work, we carefully study decentralized learning with unlabeled data  through the lens of  self-supervised learning (SSL), specifically contrastive visual representation learning. 
 We study the effectiveness of a range of contrastive learning algorithms under decentralized learning setting, on {relatively large-scale} datasets including ImageNet-100, MS-COCO, and a new real-world robotic warehouse  dataset.  Our experiments show that the \emph{decentralized} SSL (Dec-SSL) approach is {\it robust} to the heterogeneity of decentralized datasets, and learns useful representation for object classification, detection, and segmentation tasks, even when combined with the simple and standard 
 decentralized learning algorithm of  Federated Averaging (\texttt{FedAvg}).
  This robustness makes it possible to significantly reduce communication and to reduce the participation ratio of data sources with only minimal drops in performance.  Interestingly,  {using the same amount of data}, the representation learned by Dec-SSL can not only perform on par with that learned by  centralized SSL which requires communication and excessive data storage costs, but also  sometimes outperform representations extracted from 
  decentralized SL which requires extra knowledge about the data labels. Finally, we provide  theoretical insights into   understanding why data heterogeneity is less of a concern for Dec-SSL objectives,  and introduce {feature alignment and clustering techniques} to develop a new Dec-SSL algorithm that  further improves  the performance,  in the face of highly non-IID data. Our study presents positive evidence to embrace {\it unlabeled data} in decentralized learning, and we hope to provide new insights into whether and why decentralized SSL is effective and/or even advantageous.\footnote{Code is available at \url{https://github.com/liruiw/Dec-SSL}}
 \end{abstract}

\section{Introduction}

The success of machine learning hinges heavily on the access to large-scale and diverse datasets. In practice, most data are generated from different locations, devices, and embodied agents, and stored in a distributed fashion. Examples include a fleet of self-driving cars collecting a massive amount of streaming images under various road and weather conditions during everyday driving, or individuals using mobile devices to take photos of objects and scenery all over the world. 
Besides being large-scale, these datasets have two salient features: they are {\it heterogeneous} across data sources, and mostly {\it unlabeled}. For instance, images of road conditions, which are expensive to label, vary across cars driving on highways vs. rural areas, and under sunny vs. snowy weather conditions (Figure~\ref{fig:xz_distribution_vis}). 
%

Methods that can make the best use of these large-scale distributed datasets can significantly advance the performance of current machine learning algorithms and systems.
This has thus motivated a surge of research in {\it decentralized  learning/learning from decentralized data}\footnote{{Hereafter, we often use {\it decentralized learning} as a shorthand for  {\it learning from decentralized data}.}}~\citep{konevcny2016federated,hsieh2017gaia,mcmahan2017communication,kairouz2021advances,nedic2020distributed}, where usually a global model is trained on the distributed datasets using communication between the local data sources and a centralized server, {or sometimes even only among the local data sources}.   The goal is typically to reduce or eliminate the exchanges of local raw data to save communication costs and protect data privacy. 
How to mitigate the effect of {\it data heterogeneity} remains one of the most important research questions in this area~\citep{zhao2018federated,hsieh2020non,karimireddy2020scaffold,ghosh2020efficient,li2021federated}, as it can heavily downgrade the performance of decentralized learning.
Moreover, most existing decentralized learning studies focused on {\it  supervised learning} (SL) problems that require data labels \citep{mcmahan2017communication,jeong2020federated,hsieh2020non}.  Hence, it remains unclear whether and how decentralized learning can benefit from large-scale,  heterogeneous, and especially unlabeled datasets typically encountered in the real world. 
  
On the other hand, people have developed effective methods of learning purely from unlabeled data and demonstrated impressive results.
Self-supervised learning (SSL), a technique that learns {\it representations} by generating supervision signals from the data itself, has unleashed the power of unlabeled data and achieved tremendous successes for a wide range of downstream tasks in computer vision \citep{he2020momentum,chen2020simple,he2021masked}, natural language processing \citep{devlin2018bert,sarzynska2021detecting}, and embodied intelligence   \citep{sermanet2018time,florence2018dense}.
These SSL algorithms, however, are usually trained in a {\it centralized} fashion by pooling all the unlabeled data together, without accounting for the  heterogeneous  nature of the  decentralized data sources. 
{Very recently, 
there have been  
a few contemporaneous/concurrent attempts   \citep{he2021ssfl,zhuang2021collaborative,zhuang2022divergence,lu2022federated,makhija2022federated} that bridged   {unsupervised/self-supervised learning} and decentralized learning,  with focuses on {\it designing better algorithms}  that mitigate the data heterogeneity issue.  
In contrast, we revisit this new  paradigm and ask the question:  

\begin{center}
{\it 
\vspace{-7pt}
Does learning from decentralized  non-IID unlabeled data  really benefit from SSL?
\vspace{-7pt}
}
\end{center}

We focus 
on  {\it understanding} the use of SSL  in decentralized learning when handling unlabeled data. We aim to answer whether and when decentralized SSL (Dec-SSL) is effective (even combined with simple and  off-the-shelf decentralized learning algorithms, e.g., \texttt{FedAvg}  \citep{mcmahan2017communication}); what are the unique inherent properties of Dec-SSL  compared to its SL counterpart; how do the properties play a role in decentralized learning, especially with highly heterogeneous  data?   We also aim to   validate our observations on large-scale and practical datasets. 
We defer a more detailed comparison with these most related works to  \S\ref{sec:detailed_related_work}.}




{
In this paper, we show that 
unlike in decentralized (supervised) learning, data heterogeneity can be {\it less concerning} in decentralized SSL, with both empirical and theoretical evidence. This leads to more communication-efficient and robust decentralized learning schemes, which can sometimes even outperform their supervised counterpart that assumes the availability of label information.  
Among the first  studies to bridge decentralized learning and SSL, our study provides positive evidence to embrace {unlabeled data} in decentralized learning, and provides new insights into this setting. 
 We detail our contributions as follows.} 

\textbf{Contributions.}
%
{(i) We show that decentralized SSL, {specifically contrastive visual representation learning,} is a viable learning paradigm to handle 
relatively large-scale unlabeled datasets, 
even when combined with the simple {\tt{FedAvg}} algorithm.
Moreover, we also  provide both experimental evidence and theoretical insights that  decentralized SSL can be inherently {\it robust} to the  data heterogeneity   across different data sources. This allows more local updates, and can significantly improve the {\it communication efficiency}  in decentralized learning. 
(ii) We provide further empirical and theoretical evidences that even when {\it labels} are available and decentralized supervised learning (and associated representation learning) is allowed, Dec-SSL still stands out in face of highly non-IID data.}  
(iii) To further improve the performance of Dec-SSL, we design a new Dec-SSL algorithm, \texttt{FeatARC}, {by using} an iterative feature alignment and clustering procedure. Finally, we  validate our hypothesis and algorithm in practical and large-scale data and task domains, including a new real-world robotic warehouse dataset.  

\section{{Preliminaries and Overview}}  \label{sec:formulation}



Consider a decentralized learning setting with $K$ different  data sources, which  
 might correspond to different devices, machines, embodied agents, or datasets/users that can generate and store data locally. The goal  is to collaboratively solve a learning problem, by exploiting the  decentralized data from all data sources.   More specifically, 
 consider 
 each data source $k\in[K]$ has local dataset $D_k=\{x_{k,i}\}_{i=1}^{|D_k|}$, and  $x_{k,i}\in\cX\subseteq \RR^d$ are identically and independently distributed (IID) samples from probability  distribution $\cD_k$, i.e., $x_{k,i}\sim \cD_k$. Note that the distributions $\cD_k$ is in general different across  data sources $k$, yielding an overall  {\it heterogeneous} (i.e., non-IID) data distribution for the data from all the sources. Let $D=\bigcup_{k\in[K]}D_k$ denote the set of all data samples. Moreover, we are interested in situations where no label is provided alongside the data $x$. To effectively utilize the large-scale {\it unlabeled}  data, we resort to self-supervised learning approaches.

Specifically, SSL approaches extract representations from these unlabeled data, by finding an embedding function $f_w:\cX\to\RR^m$, where $w$ is the parameter of the embedding function. $z=f_w(x)$ is the representation vector that can be useful for downstream tasks, e.g., classification or segmentation.  We summarize several popular SSL approaches here that will be used later in the paper. 
\textbf{Self-supervised representation learning.}
Now consider a given data source $k\in[K]$. There are two popular  methods in the SSL community. In   contrastive learning \citep{chen2020simple,he2020momentum} specifically, a sample $x$ is used to provide supervision signals along with two generated {\it positive} {samples $x^{+}$ and $x$ (overloaded for notational simplicity)} and  (possibly multiple) {\it negative} samples $x^{-}$ sampled from the training batch.  
The goal of SSL is to find an embedding $f_w$ that makes  $x$ and $x^{+}$ close, while keeping $x$ and $x^-$s apart, if negative samples are used.  

One commonly used loss for SSL is the InfoNCE loss \citep{oord2018representation}, which has been used in popular SSL approaches as 
SimCLR \citep{chen2020simple} and MoCo \citep{he2020momentum}: 
\small
\begin{align}
\cL_k(w):= \frac{1}{|D_k|}\sum_{i=1}^{|D_k|} -\log\Bigg(\frac{\exp(-\DD(f_w(x_{k,i}), f_w(x_{k,i}^{+}))/\tau)}{{\exp(-\DD(f_w(x_{k,i}), f_w(x_{k,i}^{+}))/\tau) +\sum_j \exp(-\DD({f_w(x_{k,i}), f_w(x_{k,j}^-))/\tau)}}}\Bigg)
    \label{infonce}
    \vspace{-2pt}
\end{align}
\normalsize 
where  
$\tau>0$ is a temperature hyperparameter, $j$ is the index for negative samples, $\DD(\cdot,\cdot)$ is a distance function such as the cosine distance{, i.e., $\DD(z_1,z_2)=-\frac{z_1\cdot z_2}{\norm{z_1}\norm{z_2}}$}. Some other effective SSL approaches, such as 
 BYOL \citep{grill2020bootstrap} and SimSiam \citep{chen2021exploring}, remove the terms related to negative samples in \eqref{infonce}. These methods also add an additional function $g$, the {\it feature predictor}, which only applies to $x$ to create an asymmetry and to avoid the collapsed solutions. This usually leads to the following objective: 
$\cL_k(w):=\frac{1}{|D_k|}\sum_{i=1}^{|D_k|}\DD\big(g(f_w(x_{k,i})),f_w(x_{k,i}^{+})\big). 
    \label{simsiam}$
In our experiments, we make use of both losses and the  SSL approaches associated with them. 

\textbf{Decentralized SSL.} To exploit the heterogeneous data distributed at different locations/devices, decentralized SSL optimizes the following global objective:
\begin{align}\label{equ:obj_1}
\min_{w}~~\sum_{k\in[K]}\frac{|D_k|}{|D|}\cL_k(w),
\vspace{-2pt}
\end{align}
which can be solved using many   existing decentralized learning algorithms. For instance, {\texttt{FedAvg}}   \citep{mcmahan2017communication} is one of the most representative, easy-to-implement, and communication-efficient decentralized learning  algorithms which optimizes this objective without data-sharing among data sources. 
At each iteration $t$,  the server   first samples a set of data sources $\cM_t$ with size $|\cM_t|= \rho K$ and run $\delta$  local update steps on each of the local dataset. {  Then, each data source $k\in \cM_t$ sends back the updated local model weight  $w^{t,\delta}_{k}$  to the central server, and the server averages them to be the global model $w^{t+1}=\frac{1}{|\cM_t|}\sum_{k\in \cM_t}w^{t,\delta}_{k}$ for the next round $t+1$. The server then broadcasts the global model to each data source to reset $w^{t+1,0}_{k}$ as $w^{t+1}$. The number of local updates $(\delta)$ determines the communication efficiency (larger $\delta$ means less communication); in the experiments, we use  $E$ to denote the number of epochs of local updates (as a surrogate for $\delta$). Both $E$ and the participation rate $\rho$ are important factors that determine the efficiency of decentralized learning.} The {learned representation $f_w(x)$}  can then be used in downstream supervised learning tasks. There are many  real-world applications  of decentralized SSL, including  self-driving cars, warehouse robots, and mobile devices. A further discussion can be found in Appendix \S \ref{appendix:discussion}.    

\subsection{ Overview of Our Study }\label{sec:overview_short}

 
{
\paragraph{Terminology \& setup.} 
We separate our experiment pipeline into \textbf{representation learning} (pretraining phase) and \textbf{downstream evaluation}  (evaluation phase). Our main focus is on the  aforementioned  \textbf{Dec-SSL} approach. We use \texttt{FedAvg}  \citep{mcmahan2017communication} with SimCLR \citep{chen2020simple} as the default method.  Moreover, we will also compare with settings where the {\it label information} is available, i.e., the classical decentralized (supervised) learning, which should be more favorable for learning.  See  Figure~\ref{fig:ssl_vs_slrep} for a summary of different settings. The first setting is \textbf{Dec-SL}: we simply run \texttt{FedAvg} on the decentralized labeled data, for end-to-end classification.{Dec-SL} does not learn {\it representations} explicitly, and serves as a natural
baseline when labels are available. The second setting is {\it representation learning} from {Dec-SL}, where we train supervised learning with \texttt{FedAvg}, and then use the feature extractor network as the backbone for downstream tasks. {This way, we can also learn the representation from decentralized labeled data, and make the comparison with Dec-SSL more fair, since both  are learning features for various downstream tasks.}  We term this setting as \textbf{Dec-SLRep}.

\begin{wrapfigure}[13]{r}{0.65\textwidth} 
\centering 
  \vspace{-25pt}
\includegraphics[width=0.98\linewidth ]{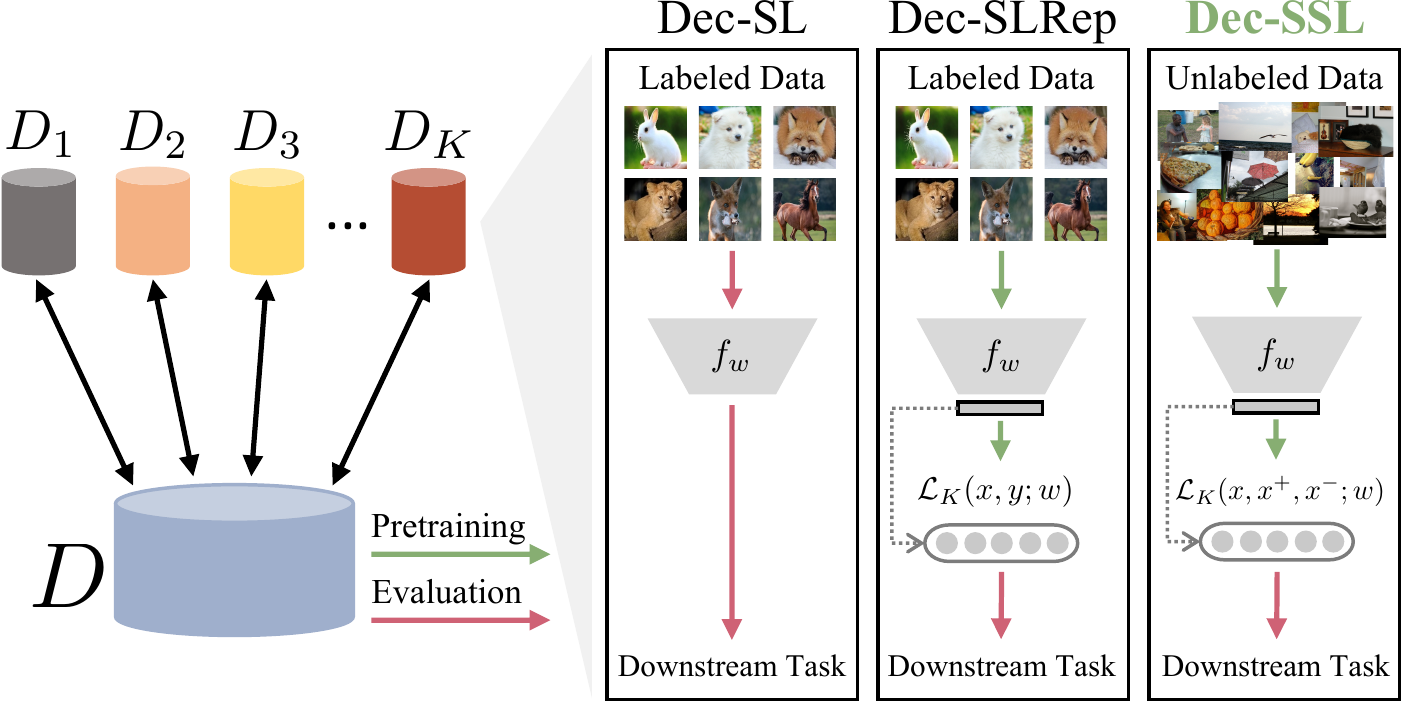} 
  \caption{\small Comparisons among Dec-SL, Dec-SLRep, and Dec-SSL. 
  } 
  \label{fig:ssl_vs_slrep}
\end{wrapfigure}


 The evaluation phase tests the representations  from Dec-SSL or Dec-SLRep. 
  We consider two protocols in the evaluation phase: \textbf{linear probing} for image classification \citep{zhang2016colorful} and \textbf{finetuning} for object detection/segmentation \citep{doersch2015unsupervised}. For classification, we train a linear classifier on top of the frozen pretrained network and evaluate the top-1 classification accuracy. For object detection/segmentation, we finetune the network by using the pretrained weights  as initialization and training in an end-to-end fashion, and then we evaluate the mean Average Precision (mAP) metric. Downstream tasks are performed on centralized train and test dataset. Please refer to Appendix \S\ref{appendix:impl} for implementation details and Table \ref{tab:experiment_setup} for experiment setups.

\textbf{Questions of interest.} 
Through extensive experiments on  large-scale datasets, and theoretical analysis in simplified settings,  we seek to answer the following questions: 
 {(i)  How well can decentralized SSL, even instantiated with the simple  \texttt{FedAvg} algorithm, rival the performance of its centralized  counterpart, and handle the non-IIDness of decentralized unlabeled data? (ii) Is there any unique and inherent  property of Dec-SSL, compared to its supervised learning counterpart; how and why may the property benefit decentralized learning, even when the label information is available? (iii) Is there a way to further improve the performance of Dec-SSL in face of highly non-IID data?  
 Our hypothesis is that SSL, whose objective is not particularly dependent on the $x$ to $y$ mappings,  
 learns a relatively {\it uniform}  representation across decentralized and  heterogeneous unlabeled datasets, thus leading to more efficient and robust decentralized learning schemes. We aim to validate this hypothesis and answer these questions in the following sections. 
}}

\section{Dec-SSL is Efficient and Robust to  Data Heterogeneity}  
\label{sec:dec_ssl_robust}

 
{We first seek to address question (i) in \S\ref{sec:overview_short} -- how well decentralized SSL performs, in face of  non-IID and decentralized unlabeled data. To this end,} 
 we first introduce {the notion of {\it data}} {\it heterogeneity} in decentralized learning, which is usually  categorized as  \textit{input heterogeneity, label distribution heterogeneity, and the heterogeneity in the relationships between the features and labels}, respectively  \citep{hsieh2020non}. We create {\it label heterogeneity}  by distributing each data source  with different 
proportion of classes; we construct {the heterogeneity} via either sampling from a Dirichlet  process with hyperparameter $\alpha$ or via skewness partitioning \citep{hsieh2020non} with hyperparameter $\beta$. We also create \textit{input heterogeneity} by leveraging the feature space of a pretrained network on the data. 
See \S \ref{appendix:data_hetereneity} for more details on how we create data heterogeneity across data sources.    
 
\subsection{Experimental observations }\label{sec:experimental_obs_robust}

\begin{figure*}[!tb] 
\centering 
\vspace{-27pt}
\includegraphics[width=1\linewidth ]{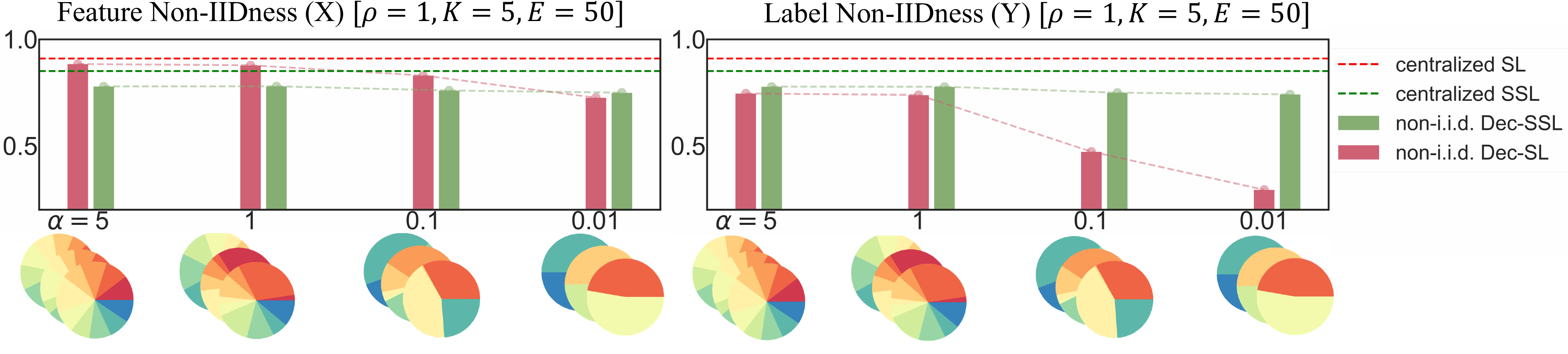} 
  \caption{{\small \textbf{ SSL objective is robust to different types of X and Y heterogeneity on the CIFAR-10 dataset.} In the pie chart below, each pie  denotes one data source, and color denotes the sample number of one source of non-IIDness (left to right, more non-IID). We observe that Dec-SSL is surprisingly  robust to the non-IIDness in both input ($X$) and label ($Y$) and also behaves closer to its centralized counterpart. Y-axis denotes  accuracy.\vspace{-20pt}}}
  \label{fig:robust_main}
  \vspace{-9pt}
\end{figure*}

\textbf{CIFAR classification under different types of non-IIDness.}\label{sec:CIFAR_noniid} In this experiment, we construct input and label non-IIDness using 5 data sources in the CIFAR-10 \citep{krizhevsky2009cifar} dataset based on the Dirichlet Process. The sources of non-IIDness are the feature clusters and labels, respectively. We control parameter $\alpha$ to create  datasets from very IID (each data source has roughly a uniform distribution over $10$ classes / $5$ feature clusters) to very non-IID  (each data source has data from $2$ classes / $1$  feature clusters). Recall that $E$ denotes the number of epochs for local updates and $\rho$ denotes the participation ratio of data sources at each round. We use $E=50$  epochs {of local updates} in this experiment, which is equivalent to around  $\delta=1000$ iterations, i.e., each local data source updates  $50$ epochs independently before averaging.  {The results are shown in Figure  \ref{fig:robust_main}.} Surprisingly, the performance of downstream classification, with representations trained using decentralized SSL, is very {\it insensitive}  to the non-IIDness across the datasets and only bears a slight performance drop.  This robustness over data non-IIDness is encouraging, and {stands in sharp contrast with most existing decentralized supervised learning algorithms, which are known to suffer from the data heterogeneity in general  \citep{hsieh2020non}. As a baseline, we consider the classical decentralized SL approach of \texttt{FedAvg}, trained over the same non-IID  data, but with label information. Indeed, the performance of decentralized SL can drop significantly as the non-IIDness increases. Finally, we note that the simple use of \texttt{FedAvg} in SSL can achieve performance comparable to the centralized SSL, showing that Dec-SSL is an effective decentralized learning scheme to handle unlabeled data.}



\textbf{Finetuning ImageNet representation for COCO detection.} 
 In this experiment, we finetune the representations learned from ImageNet to COCO detection benchmark \citep{lin2014microsoft} with the Detectron pipeline~\citep{Detectron2018}. Specifically, we use ImageNet-100 with ResNet-18 and $1\times$ training schedule for Mask R-CNN~\citep{he2017mask} with a ResNet18 FPN being the backbone. {Compared to the contemporary works  \citep{zhuang2022divergence,lu2022federated} on federated self-supervised learning, our setup is more relevant to real-world applications, as it works on larger-scale and more practical  datasets and tasks.}
 

We run Dec-SSL on ImageNet-100 dataset with $5$ data sources, and {with $E=1$ epoch of local updates}, which {corresponds to} around $\delta=500$ {local updates}, to learn the global representation {using \texttt{FedAvg}}. {On  Table \ref{tab:amazon_detection}} left, we observe that the representation from Dec-SSL almost  reaches
 the  performance of the representation from centralized SSL and improves upon  baselines that train the model from scratch{, i.e., the {\it no pretrain} row}.  This conveys that SSL can learn useful representations in decentralized settings, avoiding the heavy communication cost of centralized  learning.
 

\textbf{Decentralized SSL for real-world package segmentation.} {The issue of data heterogeneity and communication efficiency is significant for real-world applications such as those  in Amazon warehouses, whose fleets of working robots can generate millions of images per day   (see Figure~\ref{fig:amazon_distribution_vis} for an illustration). We provide details about the Amazon dataset in \S\ref{Amazon Data}.} We use data from one sample warehouse site at Amazon, and split the data based on the session ID (which is usually a sequence of days). Each decentralized learner is only allowed to  access the local data at one session, which is equivalent to the non-IID case where skewness $\beta=0$. 
We then deploy decentralized self-supervised learning on a subset of the enormous warehouse data, which has around $80000$ images with contour labels output by the Amazon work-cells. We use SimCLR with {\texttt{FedAvg}} and communication efficiency $E=1$  number of local update epochs, as the pretraining  method.     

On the right subtable of Table  \ref{tab:amazon_detection}, we compare different ways to initialize weights for finetuning, and show that the representations learned from decentralized SSL outperforms training from scratch and even matches centralized SSL on the Amazon dataset. We also experiment with finetuning segmentation task using Mask R-CNN on different fractions of the data, and show that Dec-SSL can further improve  the performance of training from scratch, when there is no as much labeled data.  


\subsection{Theoretical insights}
\label{sec:theroy}

We now provide some theoretical insights into  why the  objective of Dec-SSL leads to more robust performance in face of data heterogeneity. In particular, we analyze the property of the solutions to the local and global objectives of Dec-SSL in a simplified setting, and show that the global objective is not affected significantly by the heterogeneity of  local  datasets.  Our setup is 
inspired by the very recent work \citep{liu2021self}{, where the effect of imbalanced data in centralized SSL was studied in a simplified  setting. In particular, we generalize the centralized and $3$-way classification setting to a decentralized and $2K$-way one, carefully design the generation of data distribution across data sources, and establish analyses for both local and global objectives in decentralized SSL. We also improve some  analysis therein, and  design new metrics to characterize the performance  adapted to the decentralized setting. Due to space limitation, we include an abridged introduction here, and  defer more details to Appendix \S\ref{appendix:proof}.}
 \begin{figure*}[!tb] 
 \vspace{-27pt}
\centering 
\includegraphics[width=0.85\linewidth]{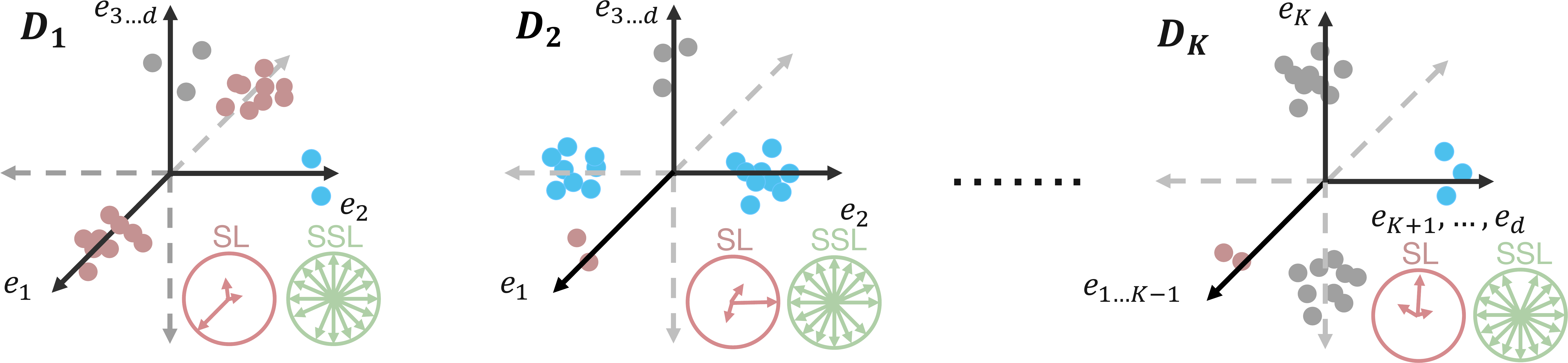}
\vspace{-6pt}
  \caption{
  \small  
   \textbf{The learned feature space of SSL is more insensitive to heterogeneity under the linear settings.} In  \S \ref{sec:theroy}, we consider a decentralized learning setting where each local dataset has a skewed distribution with most data points (each color is a class) concentrated on one axis. Each basis vector inside the sphere denotes how well it is represented in the learned subspace. For contrastive objectives, the learned feature space (green sphere) of the local model is more uniform and close to the global model. On the other hand, the SL objective (red sphere) tends to overfit to local dataset, and the learned feature spaces become heterogeneous.}
   \vspace{-3pt}  
  \label{fig:theory_plot}
  
\end{figure*}
\begin{table*}[t]
 
\centering
{\centering 
\footnotesize \setlength\tabcolsep{4pt}
\begin{tabular}[t]{c|cc} 
\hline 
 ImageNet-100 &
  \multicolumn{2}{c}{MS-COCO  }        \\  
  Pretrain   & 
  \multicolumn{1}{c}{ \apbbox{~}}  & \multicolumn{1}{c}{\apmask{~}}
\\ \hline
 \color{Gray}no pretrain   &  \color{Gray} {20.5} & \color{Gray}{19.4}   \\ 
Central-SLRep    &   21.2\cgaphl{+}{0.7} &  20.1\cgaphl{+}{0.7}     \\ 
Central-SSL    &  {23.2}\cgaphl{+}{2.7} &  {22.1}\cgaphl{+}{2.7}        \\
\hline
Dec-SLRep   &   {19.8}\cgaphl{-}{0.7} &  {19.7 }\cgaphl{+}{0.3}    \\
Dec-SSL   &  {22.1}\cgaphl{+}{1.6} &   {20.7} \cgaphl{+}{1.3}
\\
\hline
\end{tabular}  
\footnotesize \qquad\quad
\begin{tabular}[t]{c|ccc}
\hline 
\multicolumn{0}{c|}{Amazon}   &\multicolumn{3}{c}{Amazon (\apmask{~})}
   \\  
  \multicolumn{0}{c|}{Pretrain}   & \multicolumn{0}{c}{$100\%$} & \multicolumn{0}{c}{$10\%$}& \multicolumn{0}{c}{ $1\%$ }          \\ \hline
  
\color{Gray}no pretrain  &  \color{Gray}60.8  & \color{Gray}59.2  & \color{Gray}47.0 \\
Central-SSL  & 61.6\cgaphl{+}{0.8}  & {60.4}\cgaphl{+}{1.2}  & {49.5}\cgaphl{+}{2.5} \\
Dec-SSL  &   {61.2}\cgaphl{+}{0.4} & 60.1\cgaphl{+}{0.9}  & 48.8\cgaphl{+}{1.8}     \\
\hline
\end{tabular}}
\vspace{-7pt}
\caption{
\small
\textbf{Left: Object detection and semantic segmentation finetuned on COCO:} The model is pretrained on ImageNet-100~\citep{tian2020contrastive} dataset and then finetune on MS-COCO with metrics bounding-box mAP (\apbbox{~}) and mask mAP (\apmask{~}). 
  \textbf{Right: Finetuning results on the Amazon package segmentation dataset with representations pretrained on the Amazon dataset.} We observe that Dec-SSL reaches similar performance (\apmask{~}) as centralized SSL and also outperforms training from scratch. Note that $100\%,10\%,1\%$ denote the portion of the data used for finetuning.\vspace{-12pt} }
\label{tab:amazon_detection} 
\end{table*}

{\textbf{Setup.} Consider a Dec-SSL problem with $K$ data sources. 
 Similar to the SimSiam approach, we first augment $x$, an anchor sample from the dataset, by sampling $\xi,\xi'\sim\cN(0,I)$ IID from the Gaussian distribution. Consider the linear embedding function $f_w(x)=wx$, where $w\in\RR^{m\times d}$ and $m\geq 2K$. The SSL objective for data source $k$ is given by
\#  \label{equ:local_obj_theory_abridged}
\cL_k(w):=
-\hat{\EE}\big[(w(x_{k,i}+\xi_{k,i}))^\top(w(x_{k,i}+\xi_{k,i}'))\big]+\frac{1}{2}\|w^\top w\|_F^2, 
 \#
where $\hat{\EE}$ is taken expectation over the empirical  dataset $x_{k,i}\sim D_k$, and the randomness of  $\xi_{k,i}$ and $\xi_{k,i}'$.  
Moreover, recall the global objective is given in \eqref{equ:obj_1}. Note that \eqref{equ:local_obj_theory_abridged} instantiates SimSiam loss with the negative inner-product $\la a, b\rangle$ as the distance function $\DD(a,b)$ and 
no feature predictor, and with a regularization term for  mathematical  tractability, as in \citet{liu2021self}.   

\textbf{Data heterogeneity.}
The $K$ data sources collaboratively solve \eqref{equ:obj_1} to learn a representation for a $2K$-way classification task. 
The $K$  local datasets are generated in a way that for each fixed $k\in[K]$, 
the labels are skewed in that data from classes $2k-1$  and $2k$ constitute the majority of the data, while other classes are rare, or even unseen. 
More details on the specifications of data heterogeneity 
can be found in \S\ref{sec:ssl_proof}. 
We visualize the heterogeneity of the data distributions in Figure \ref{fig:theory_plot}. 
}

{To compare the representations learned across data sources and that learned from jointly solving \eqref{equ:obj_1}, we introduce the following definition on the representability of the representation space.}

\begin{definition}[Representability vector] 
Let $\cS\subseteq\RR^d$ be the subspace spanned by the rows of the  learned feature matrix  $w\in\RR^{m\times d}$, where the embedding function $f_w(x)=wx$.  The {\it representability} of $\cS$ is defined as a vector  $\br=[r_1,\cdots,r_d]^\top\in\RR^{d}$ , such that $r_i=\|\Pi_{\cS}(e_i)\|_2^2$ for $i\in[d]$, where $\Pi_{\cS}(e_i)\in\RR^d$ is the projection of standard basis $e_i$ onto $\cS$, and thus $r_i=\sum_{j=1}^{s}\la e_i, v_j\rangle^2$ where 
$s=\text{dim}(\cS)$ and  $\{v_1,\cdots,v_s\}$ is a set of orthonormal bases for $\cS$.  
\end{definition}
 
The intuition of this definition is that a good feature space should have the property that many standard unit bases among  $e_1,\cdots,e_d$,  which can be used to represent any vectors in $\RR^d$, can be represented well by the feature space, i.e., have large projections onto it. Note that as a vector,  $\bm{r}$ provides a quantitative way to compare the representability of two feature spaces across  different directions (i.e., different unit basis).  In the following theorem, we compare the representability learned by local objectives and the global one, for Dec-SSL.


\begin{theorem}[Representability of  local v.s. global objectives for Dec-SSL]\label{thm:main} 
For decentralized SSL in the setting described above, with high probability, the representability vector learned from any local objective of source $k$, denoted by $\br^k=[r_1^k,\cdots,r_d^k]^\top$, satisfies that $1-O(d^{-4/5})\leq r^k_i\leq 1$ for all $i\in[K]\setminus{\{k\}}$. 
Moreover, the representability vector learned from the global objective, denoted by $\bar{\br}=[\bar{r}_1,\cdots,\bar{r}_d]^\top$,  satisfies that $1-O(d^{-4/5})\leq \bar{r}_i\leq 1$ for all  $i\in[K]$.   \vspace{-2pt} 
\end{theorem}

Theorem \ref{thm:main} states that the feature spaces learned from local SSL objectives are relatively {\it uniform}, in the sense that for the $K$ basis directions $e_1,\cdots,e_K$ that generate the data, any two data sources have similar representability in {\it all of them but two}  directions, especially when the dimension $d$ of the data  is large.  Furthermore, when  solving the global objective \eqref{equ:obj_1}, the learned representation is also uniform, and its representability  differs {\it at most one} direction from that of each local data source. Note that the results hold with highly heterogeneous data across data sources. In other words, Dec-SSL is not affected significantly by the non-IIDness of the data, justifying the empirical observations in \S\ref{sec:experimental_obs_robust}. Illustration of the results can also be found in Figure \ref{fig:theory_plot}.


\textbf{Intuition \& implication.} The main intuition behind  Theorem \ref{thm:main} is that, the objective of SSL is not {\it biased} by the heterogeneous distribution of labels at each local dataset, and tends to learn uniform representations. Related arguments have also been made in the recent works on the theoretical understanding contrastive learning/SSL \citep{wang2020understanding,liu2021self}.  In the decentralized setting, this insensitivity to data heterogeneity becomes even more relevant, as it potentially allows each local data source to perform much more local updates, without drifting the iterates significantly. This enables more communication-efficient decentralized learning schemes, in contrast to most existing ones that are vulnerable to data non-IIDness. We validate these points next. 


\section{Dec-SSL Can be Favorable Even 
When Labels are Available
 }\label{sec:ssl_vs_sl_constraints}  

 \begin{figure*}[!tb]
 \vspace{-28pt}
\centering 
\includegraphics[width=0.9\linewidth ]{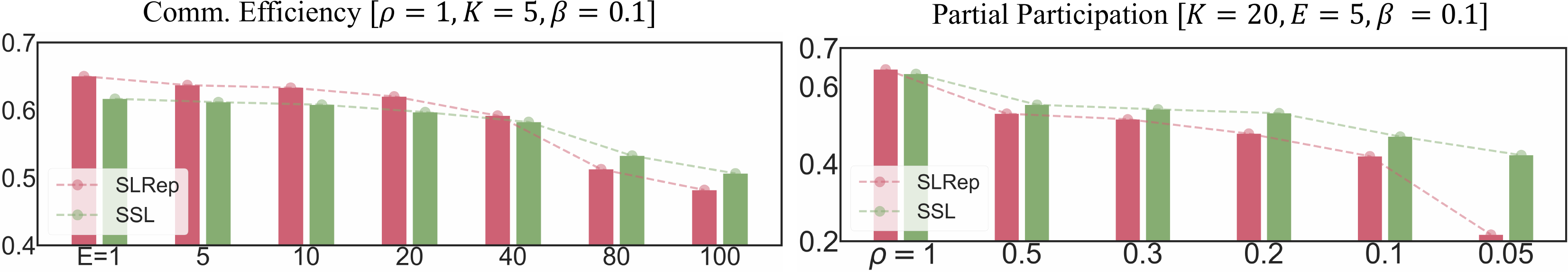} 
  \caption{\small  \textbf{Dec-SSL performance on ImageNet-100 dataset}. Compared to supervised learning, we observe that under non-IID settings, decentralized SSL can perform better under communication constraints (left) and partial participation constraints (right).\vspace{-15pt}} 
  \label{fig:ImageNet}
\end{figure*}
 
{We here seek to address question  (ii) in \S\ref{sec:overview_short} -- how does the unique property of Dec-SSL, such as the robustness to data heterogeneity, benefit decentralized learning?  {While lack of labels seems   a limitation, we show that this might not be the case in decentralized learning with heterogeneous data. First, it is known that decentralized SL in general
 performs poorly when the data is highly heterogeneous \citep{zhao2018federated,hsieh2020non}. Further, even in the decentralized representation learning setting when labels are available,  Dec-SSL still stands out in face of highly non-IID data.}
}  

{
To make a fair comparison, we mainly compare Dec-SSL with Dec-SLRep (recall the definition in \S\ref{sec:overview_short}), which are both decentralized {\it representation learning}  approaches. We defer the comparison with Dec-SL to  Appendix \S\ref{appendix:additional_exp}.  We conduct experiments on both ImageNet and CIFAR-10 datasets,  and evaluate the performance of the learned representations in terms of the variations of  two commonly used metrics in decentralized learning  -- the number of local updates epochs $E$, and the participation ratio of data sources $\rho$. We observe consistently that Dec-SSL indeed outperforms Dec-SLRep in learning representations in terms of communication efficiency and participation ratio, especially with highly non-IID data. We remark that such observations are also consistent with those on object detection and semantic segmentation given in Table \ref{tab:amazon_detection}. 
}

\subsection{Experimental observations}\label{sec:experimental_obs_sl_vs_ssl}
In this experiment, we   train and evaluate the feature backbone on ImageNet-100 in a decentralized setting. We create non-IIDness across the local datasets based on label skewness and use $\beta=0.1$ (each data source has only $10\%$ of its data coming from the uniform class distributions). 

 \begin{figure}
\vspace{-25pt}
\begin{floatrow}
\ffigbox{%
\includegraphics[width=0.93\linewidth]{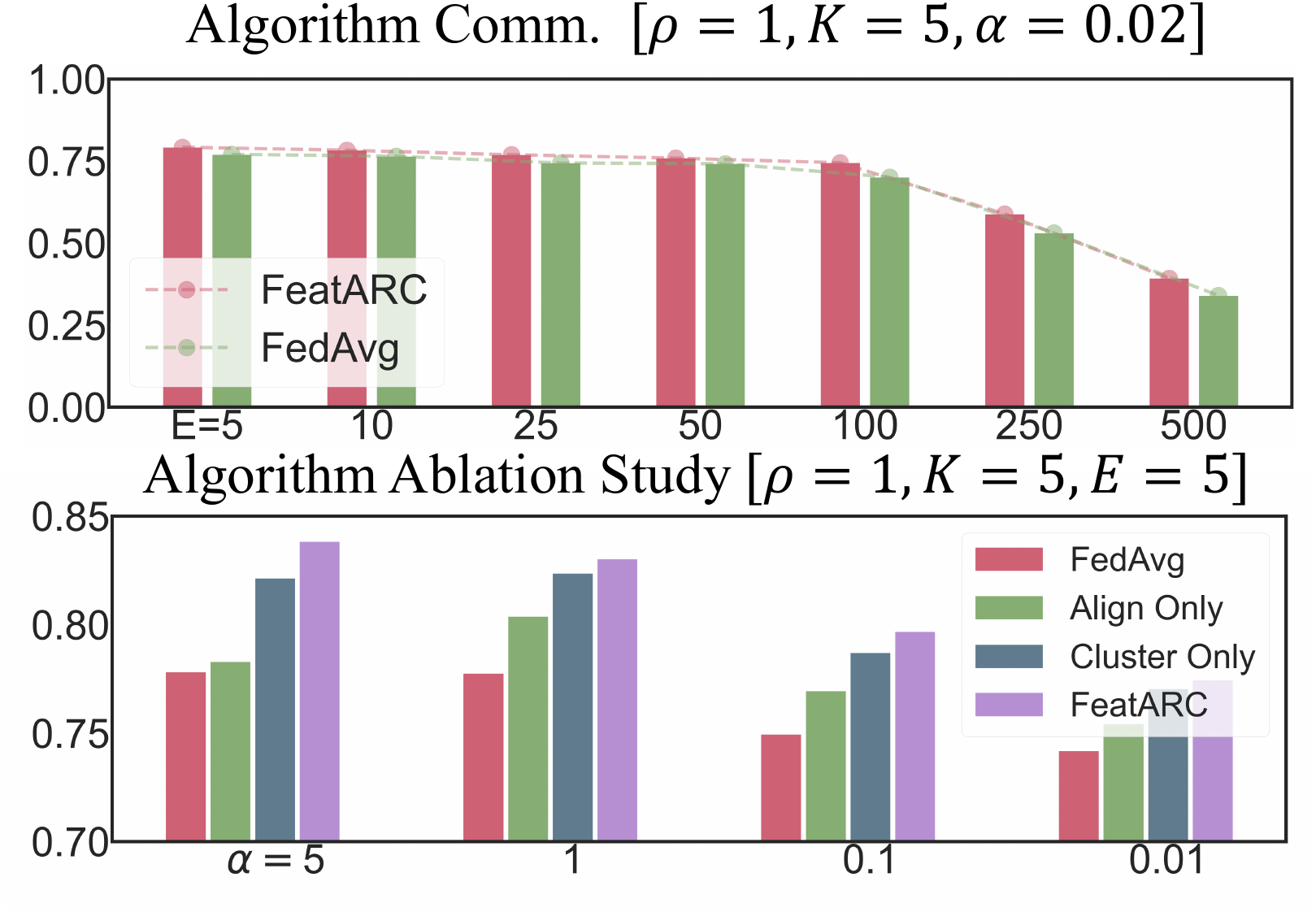} 
\vspace{-15pt}
}{%
  \caption{\small  \textbf{ Ablation study on the \texttt{FeatARC} algorithm.} We observe that under non-IIDness and communication constraints, \texttt{FeatARC} outperforms the baseline variants of the algorithm and \texttt{FedAvg}.  
  }\label{fig:algo_ablation}
}
\capbtabbox{%
{ 
\scriptsize\begin{tabular}[t]{c|ccc}
\hline 
\multicolumn{2}{c|}{Method / Setting} & 
  \multicolumn{0}{c}{IID} & \multicolumn{0}{c}{non-IID}   \\ \hline
\multicolumn{2}{c|}{FURL \citep{zhang2020federated}} & 71.25 &  68.01 \\
 \multicolumn{2}{c|}{EMA \citep{zhuang2022divergence}} &  86.26 & 83.34 \\
 \multicolumn{2}{c|}{Per-SSFL \citep{he2021ssfl}} & N/A  & 83.10  \\
 \multicolumn{2}{c|}{FEDU  \citep{zhuang2021collaborative}} & 83.96 & 80.52\\
 \multicolumn{2}{c|}{{\texttt{FeatARC}} (Ours)}  & \textbf{86.74} & \textbf{84.63}  \\
\hline
\hline 
\multicolumn{0}{c|}{CIFAR-100}      &\multicolumn{3}{c}{CIFAR-10}
  \\ 
 \multicolumn{0}{c|}{ Pretrain}  & \multicolumn{0}{c}{ 100$\%$} & \multicolumn{0}{c}{ 10$\%$} &
  \multicolumn{0}{c}{ 1$\%$} \\ 
  \hline
\color{Gray}no pretrain &  \color{Gray}0.31 &  \color{Gray}0.27 &\color{Gray}0.25  \\
Dec-SLRep IID    & 0.65 & 0.60 & 0.47 \\
Dec-SSL IID        &\textbf{0.71} & \textbf{0.67} & \textbf{0.57}  \\
\hline
Dec-SLRep Non-IID   & {0.43} & {0.35} & {0.32} \\
Dec-SSL Non-IID  & \textbf{0.70} & \textbf{0.66} & \textbf{0.57} \\
\hline
\end{tabular}}
}{%
  \caption{\small {\bf {Top). Algorithm performance  comparison.}}  
  {\bf Bottom). CIFAR-10 Linear probing on the representation of CIFAR-100.}
    Our algorithm  surpasses previous works on federated SSL both in the IID and non-IID settings.  
  }\label{tab:cifar100tocifar10}
}
\end{floatrow}
\vspace{-14pt}  
\end{figure}

\textbf{Communication efficiency under high non-IIDness.} In Figure~\ref{fig:ImageNet}, we show that under the non-IID scenario,  averaging weights with an infrequent communication schedule causes less trouble to Dec-SSL than to Dec-SLRep. In {\texttt{FedAvg}}, the idea of averaging weights after multiple epochs might sound sub-optimal, but we notice that  decentralized SSL is very robust with respect to this parameter. Intuitively, the robustness of Dec-SSL  allows each local model to drift longer, leading to a lower communication frequency for decentralized learning.



\textbf{Participation ratio under high non-IIDness.} In this experiment, we split ImageNet-100 into 20 data sources and use local update $E=5$ epochs. We measure the performance of decentralized learning algorithms with respect to the participation ratio of data sources at each round. For instance, when $\rho=1$, at each round,  all data sources update their local weights and upload to the server, while $\rho=0.05$ means that each round a single random data source is selected for update. On the right of Figure~\ref{fig:ImageNet}, we show that with non-IID data, the convergence of Dec-SSL is more stable to less participants compared to Dec-SLRep. {This allows more efficient decentralized learning, especially when deployed with extremely  large number of data sources and unstable communication channels.} 

\subsection{Theoretical insights}\label{sec:theory_2} 


To shed light on the above  observations, 
 we provide analysis for the feature spaces learned by the local objective of Dec-SLRep, under the same setup as in \S\ref{sec:theroy}. For Dec-SLRep and each data source $k$, we consider learning a two-layer linear network $g_{u_k,v_k}(x):=v_k u_kx$ as classifier, where $u_k\in\RR^{m\times d}$ and $v_k\in\RR^{c\times m}$, and use $u_k x$ as the learned representation for downstream tasks.   The network is learned by minimizing  $\|(u_k)^\top u_k\|_F^2+\|(v_k)^\top v_k\|_F^2$ subject to the margin constraint that $[g_{u_k,v_k}(x)]_y \geq [g_{u_k,v_k}(x)]_{y'}+1$ for all data $(x,y)$ in the local dataset $k$ with all $y'\neq y$. 
. We now have the following proposition on the representations learned by  Dec-SLRep across data sources.




 \begin{proposition}[Representations learned by Dec-SLRep across heterogeneous data sources] 
 \label{prop:result_sl_main}
With high probability, the features   $u_k=[u_{k,1},\cdots,u_{k,m}]^\top\in \mathbb{R}^{ m \times d}$ learned from the local dataset $D_k$ satisfies that  $
\sum_{i=1}^m\langle u_{k,i}, e_{j} \rangle^2\le O(d^{-\frac{1}{10}})$, 
 for $j\in[K]\setminus \{k\}$; while $\sum_{i=1}^m\langle u_{k,i}, e_{k} \rangle^2\ge 1-O(d^{-\frac{1}{20}})$. 
In other words, the correlation between the learned features in $w_k$ and $e_j$ is small for all $j\in[K]\setminus \{k\}$, while the correlation between the features and $e_k$ is large. 
\end{proposition}  

The proposition suggests that  the feature spaces learned by Dec-SLRep differ significantly across local data sources, given the highly heterogeneous data. More specifically, we show that most of the unit bases in $\{e_1,\cdots,e_K\}$ have small correlations with the features learned at each local data source, while these feature spaces themselves vary significantly across data sources. The unit bases that are not learned might be significant for various other downstream tasks, making the learned representations less favorable. This heterogeneity among local solutions is not in favor of  {\it local updates}, as too many local updates would drift the iterates towards its local solution, and the iterates would become too far away from each other, hurting the convergence  of decentralized learning. Hence, compared with the Dec-SSL case and  Theorem \ref{thm:main},   Dec-SLRep can be less robust to data heterogeneity and less  communication-efficient.  {We note that the advantage of Dec-SSL  does not {come from {\it using more data}}, 
since we use exactly the same data for training Dec-SLRep and Dec-SSL. The intuition is also  illustrated in Figure \ref{fig:theory_plot}.} {Finally, we remark that the {\it uniformity} of features, which is believed to be the key to better transfer performance in SSL  \citep{wang2020understanding,caron2020unsupervised}, is not always preferred given  {\it specific} learning tasks   \citep{burgess2018understanding}.}

\section{Our Algorithm --  {\tt FeatARC} (Feature Alignment and  {Clustering})} \label{sec:algorithm_main}

 Although Dec-SSL tends to learn relatively uniform features that are robust across datasets, the uniformity itself might not imply the  alignment of features across datasets: the  representation network from different local data sources can still map the same data point to different regions in the feature space. This  misalignment  becomes more significant when the  data is highly non-IID and can have an adverse effect on the model aggregation process in  decentralized learning \citep{zhang2020federated}.  {To mitigate this issue and address question (iii) in \S\ref{sec:overview_short}, we propose to} use the same feature distance loss as an auxiliary local objective to align the local models with the global model. 
  The alignment between two features is defined as the  negative  cosine distance metric $\DD(z_1,z_2)=-\frac{z_1\cdot z_2}{\norm{z_1}\norm{z_2}}$.
  

  {To further improve the Dec-SSL algorithm, we propose to learn {\it multiple models} using clustering-based approach. In particular, instead of learning a single global model as in \eqref{equ:obj_1}, we learn $C$ models and separate the $K$ data sources into $C$ clusters. The update of $C$ models and the assignment of data sources to $C$ clusters are conducted alternatively.  When $C=K$, the algorithm reduces to learning $K$ local models; when $C=1$, it reduces to learning a single global one.  {The clustering approach intuitively learns multiple models to interpolate the performance  between learning a {\it single global} model and $K$ {\it local} models, thus achieving a good bias-variance tradeoff when testing on each local dataset \citep{mansour2020three,ghosh2020efficient}}.  However, unlike the supervised learning case,  we do not use the loss of the decentralized learning (i.e., \eqref{infonce}) as the metric for  clustering. This is because for contrastive learning, it has been observed that the SSL loss  might not be indicative enough for the performance of the representation on downstream tasks \citep{robinson2021can}. Hence, we here again use  the feature alignment distance $\DD(\cdot,\cdot)$ as the metric for clustering. 
  }
 
 
 
 {We adopt the alignment regularization and clustering techniques, and developed a new Dec-SSL algorithm \texttt{FeatARC},  summarized in Algorithm \ref{alg:farc} and Algorithm \ref{alg:alignment_loss}} in Appendix.  {We show the performance of \texttt{FeatARC} in Figure \ref{fig:algo_ablation}, in comparison with different baselines including \texttt{FedAvg}, under different levels of data heterogeneity and communication frequency. It is shown that \texttt{FeatARC} outperforms the baselines consistently, including the variants that only uses alignment (``Align Only'') or clustering (``Cluster Only''). Moreover, on the 
top of Table  \ref{tab:cifar100tocifar10}, we show that \texttt{FeatARC} also outperforms other recent decentralized self-supervised learning algorithms on  CIFAR-10 dataset.}
 



\section{{Extensions}}
In this section, we discuss a few extended experiments of our framework. Please see Appendix \S\ref{appendix:additional_exp} for a thorough set of experiments and ablation studies with visualizations.

 \subsection{Fully decentralized case and  different network topology}

We conduct experiments on the {\it fully decentralized} learning in Appendix \S\ref{sec:full_dec_append},  where the local data sources are only allowed to communicate with their neighbors over a peer-to-peer network, without a centralized server. In short, most observations we had regarding Dec-SSL in the setting with a centralized server still hold, even under several different network topologies. This aligns with our theoretical insight provided in Section \ref{sec:dec_ssl_robust}, which came from the benign properties of the {\it solution} to the Dec-SSL {\it objective}, instead  of the properties of {\it specific algorithms} (averaging the iterates via a star or other network topologies) that achieves the solution.

\subsection{Extremely heterogeneous case  for decentralized learning}
In Figure~\ref{fig:extre_case}, we show that even in the extremely heterogeneous case where each local source only owns \emph{one}  class, the Dec-SSL framework is still robust to the non-IIDness of the data. This also holds true when we scale to more clients, as shown  in Figure~\ref{fig:more_client}. The Dec-SSL objective would not be {\it biased} by the highly heterogeneous  class labels at each local dataset, while the Dec-SL objective could be   biased by it. This is also consistent with our theoretical insights in Section \S\ref{sec:theroy} and the key reason for the success of Dec-SSL is that, despite only having one single class, the information of features obtained from local datasets may still be useful for the jointly classifying of  all the  classes. 

\subsection{Comparison of FeatARC with other algorithms}
We also compare our algorithm with the Dec-SSL algorithms that are combined with other federated learning algorithms, including  \cite{li2020federated} (FedProx) and \cite{li2020fedbn} (FedBN).  In Figure~\ref{fig:fedprox} (Left), we show that our proposed {\tt FeatARC} can outperform these two baselines.

\section{Conclusion}
We propose the framework of  decentralized SSL that learns representations from non-IID unlabeled data and conduct an empirical study on the robustness of Dec-SSL to different types of heterogeneity, communication constraints, and  participation rates of data sources. We also provide   findings and theoretical analyses of Dec-SSL compared to its supervised learning counterpart, as well as developing a new algorithm to further address the high heterogeneity in decentralized datasets.  

\paragraph{Acknowledgement.} 
This work is supported in part by Amazon.com Services LLC, PO2D-06310236 and Defense Science \& Technology Agency, DST00OECI20300823. L.W. was supported by the MIT EECS Xianhong Wu Graduate Fellowship. K.Z. also acknowledges  support  from Simons-Berkeley Research Fellowship. We thank MIT Supercloud for providing compute resources. The authors would like to thank many helpful discussions from Phillip Isola at MIT and Andrew Marchese at Amazon. 
\bibliography{references}
\bibliographystyle{iclr2023_conference}
\normalsize 

\newpage
\appendix
\onecolumn



~\\
\centerline{{\fontsize{13.5}{13.5}\selectfont \textbf{Supplementary Materials for {``Does Learning from Decentralized}}}}

\vspace{6pt}
\centerline{\fontsize{13.5}{13.5}\selectfont \textbf{{Non-IID
Unlabeled 
Data Benefit from Self Supervision?''}}}  
\vspace{10pt}
  
\tableofcontents 

\newpage
 \setlength{\abovedisplayskip}{4pt}
\setlength{\belowdisplayskip}{4pt}
\titlespacing{\section}{0pt}{0pt}{0pt}
\titlespacing{\subsection}{0pt}{0pt}{0pt}
\titlespacing{\subsubsection}{0pt}{0pt}{0pt}

\section{Detailed Related Work}\label{sec:detailed_related_work}

We here provide a more detailed  review of the literature.  

 
\paragraph{Self-supervised learning.} 
Self-supervised learning   aims to learn useful representations from data without human annotations. 
A breadth of methods has been proposed such as colorization~\citep{larsson2016learning}, impainting~\citep{pathak2016context}, and denoising autoncoder~\citep{vincent2008extracting}. One promising approach is contrastive learning,  where the core idea is to find a representation space that makes positive pairs close and negative pairs apart.  \citep{oord2018representation,wu2018unsupervised,tian2020contrastive,chen2020simple,he2020momentum,chen2021exploring,grill2020bootstrap,caron2020unsupervised} used a self-supervised pretraining objective for transformation-invariant representation, and demonstrated good performance on standard datasets such as ImagetNet classification  \citep{krizhevsky2012imagenet},  COCO detection and segmentation \citep{lin2014microsoft}, and uncurated dataset \citep{caron2019unsupervised}.  Despite that no label is needed, the representation learned by SSL has been shown to be  robust to distribution shift  \citep{liu2021self},  generally applicable in embodied agent tasks  \citep{florence2018dense,hendrycks2019using}, and adapt quickly in low-data regime \citep{tian2020rethinking}.  More importantly, SSL models have been deployed in many large systems (foundation models \citep{bommasani2021opportunities}) in NLP \citep{brown2020language} and intersection of language and vision \citep{radford2021learning,jia2021scaling}.
Other significant examples of SSL include masked auto-encoding in language \citep{devlin2018bert} and vision \citep{he2021masked}.
There has also been a growing  literature on the theoretical  understandings of SSL, with representative examples \citep{arora2019theoretical,lee2021predicting,tosh2021contrastive,haochen2021provable}. Very recently,  \citep{liu2021self}  observed that self-supervised learning is more robust to dataset imbalance, more specifically, the data label  imbalance. Interestingly,  their observations are aligned with ours with decentralized heterogeneous data (though we focus on  not only  the label skewness among data sources), and their analysis also provide  important insights into our observations.  While all these works consider SSL in a centralized setting, our goal is to further understand and unlock the power of  SSL in a decentralized setting, a  practical while  relatively underexplored  one where {\it large-scale unlabeled} data is more relevant.     

\paragraph{Decentralized machine learning.}

With massive amounts of data generated in a distributed fashion, decentralized learning has achieved increasing attention in the literature \citep{konevcny2016federated,lian2017can,mcmahan2017communication,hsieh2017gaia,hsieh2020non,karimireddy2020scaffold,kairouz2021advances,nedic2020distributed}, where a global model is trained  over distributed data sources, addressing research questions  on  communication-efficiency (between the worker and the server or among workers) and privacy of the data. In addition, addressing the  {\it heterogeneity/non-IIDness} of data distributions across sources has been,  and remains to be the most important and challenging  research question in the area  \citep{zhao2018federated,hsieh2020non,karimireddy2020scaffold,ghosh2020efficient,li2020federated,li2020fedbn,li2021federated}. Most existing decentralized learning studies have  extensively focused  on {\it supervised learning} setting, where the labels of the data samples are required.  

\paragraph{Federated unsupervised/self-supervised learning.} 
To the best of our knowledge, there have only been 
a few contemporaneous/concurrent attempts   \citep{zhang2020federated,he2021ssfl,zhuang2021collaborative,zhuang2022divergence,lu2022federated,makhija2022federated} 
that bridged {\it unsupervised/self-supervised learning} with unlabeled data and decentralized learning,  
more specifically  federated learning  (FL), 
and proposed various  algorithms to {\it mitigate} the effect of data  heterogeneity. In particular, the works \cite{he2021ssfl,zhuang2021collaborative,zhuang2022divergence,lu2022federated,makhija2022federated} are closest to ours. \cite{he2021ssfl}, also motivated by the label-deficiency issue in federated learning, developed a series of self-supervised FL algorithms that incorporated the advances of supervised FL, especially those algorithms  with {\it personalization}, to handle the heterogeneity in data. \cite{zhuang2021collaborative} developed unsupervised representation learning algorithms from unlabeled data, mainly with the motivation of privacy-preserving, by designing  communication protocol and divergence-aware predictor update rules that are {\it specific to} Siamese architecture. \cite{zhuang2022divergence} further improved the results by generalizing to other SSL approaches,  proposing a new divergence-aware update rule, and ablating on how the  components of these SSL approaches affect the performance.  Later, \cite{lu2022federated} also aimed to address the data-deficiency issue in FL, by training a {\it modified} model using {\it supervised} FL over the {\it surrogate labeled}  data transformed from the unlabeled ones. The transformation requires knowledge of the class priors at each data source, and the approach is not relevant to self-supervised/contrastive learning, the focus of our paper.   Finally, \cite{makhija2022federated} proposed a self-supervised federated learning algorithm to handle the heterogeneity in data, by adding a proximal term that measures the distance between the local representations and those obtained on other clients in the local objective. The algorithm requires the server to directly  access the unlabeled datasets, and also requires some {\it datasets} for representation alignment to be transmitted between the server and the clients. 

\paragraph{Our focus.} 
To be specific, our focus is {\it not} on 
finding better algorithms to handle/mitigate data heterogeneity in decentralized learning with unlabeled data,  but on {\it understanding} the use of self-supervised learning approaches, in particular contrastive learning, in decentralized learning -- whether and when decentralized SSL is effective and/or even advantageous (even combined with simple and off-the-shelf decentralized learning algorithms, e.g., \texttt{FedAvg}); what are the unique and inherent properties of decentralized SSL (compared to its SL counterpart); how may  the properties play a role in decentralized learning (especially with highly heterogeneous unlabeled data)?  Moreover, except \cite{lu2022federated,makhija2022federated}, 
which contained some convergence analysis  for the algorithms they developed, 
 these contemporaneous  works usually did not provide theoretical insights 
 about why decentralized SSL is used to handle decentralized unlabeled data, and when it is effective/advantageous (even sometimes the labels are available). Finally, our goal is to advocate the Dec-SSL {\it framework}, and the approach is not specific to certain network architecture (as e.g., \cite{zhuang2021collaborative,zhuang2022divergence}), and does not require transmitting datasets (e.g., \cite{makhija2022federated}). Finally,   our empirical observations are thoroughly  verified  on {\it larger-scale} datasets compared to these works, e.g., ImageNet, MS-COCO, and real-world robotic warehouse datasets, which are more relevant to practical applications. 
\begin{table*}[!t] 
\centering
{\centering \begin{tabular}{c|ccccccc}
\hline 
\multicolumn{0}{c|}{  Experiment}   &
  \multicolumn{0}{c}{ Pretrain} &  \multicolumn{0}{c}{ $K$} &  \multicolumn{0}{c}{$E$} & \multicolumn{0}{c}{$\rho$} & \multicolumn{0}{c}{$\alpha$} & \multicolumn{0}{c}{$\beta$} & \multicolumn{0}{c}{ Evaluation }   \\ \hline
 Figure \ref{fig:robust_main} & CIFAR-10  & 5 & 50 &1 & * & $-$ & CIFAR-10   \\
Table \ref{tab:amazon_detection} Left & ImageNet-100 & 5 & 1&  1 & 0.2 & $-$   & MS-COCO   \\
Table \ref{tab:amazon_detection} Right & Amazon   & 5 & 1 & 1 & $-$ & 0 & Amazon  \\
Figure \ref{fig:ImageNet} Left & ImageNet-100 & 5 & * & 1 & $-$ & 0.1  &ImageNet-100   \\
Figure \ref{fig:ImageNet} Right & ImageNet-100 & 20 & 5 & * & $-$ & 0.1  &ImageNet-100   \\
Figure \ref{fig:algo_ablation} Left Top & CIFAR-10 & 5 & * & 1 & 0.02 & $-$  & CIFAR-10  \\
Figure \ref{fig:algo_ablation} Left Bottom& CIFAR-10 & 5 & 5 & 1 & * & $-$  & CIFAR-10  \\
Table \ref{tab:cifar100tocifar10} Top & CIFAR-10 & 5 & 5 & 1 & $-$ & $-$  &CIFAR-10   \\
Table \ref{tab:cifar100tocifar10} Bottom & CIFAR-100 & 5  & 5& 1 & $-$ & $-$   & CIFAR10  \\
Figure \ref{fig:ssl_ablation} (a,b) & CIFAR-10 & 5 & * & 1 & 0.02 & $-$  &CIFAR-10    \\
Figure \ref{fig:ssl_ablation} (c) & CIFAR-10 & 5 & 50 & * & 0.02 & $-$  &CIFAR-10    \\
Figure \ref{fig:ssl_ablation} (d) & CIFAR-10 & 5 & * & 1 & 0.02 & $-$  &CIFAR-10    \\
Figure \ref{fig:ssl_ablation} (e) & CIFAR-100 & 5 & * & 1 & $-$ & 0.1  &CIFAR-100    \\
Figure \ref{fig:ssl_ablation} (f) & TinyImageNet & 5 & * & 1 & $-$ & 0.1  & TinyImageNet    \\
Figure \ref{fig:ssl_ablation} (g) & CIFAR-10 & 5 & 50 & 1 & * & $-$  &CIFAR-10    \\
Figure \ref{fig:ssl_ablation} (h) & CIFAR-10 & 5 & 50 & 1 &$-$ & *  &CIFAR-10    \\
Figure \ref{fig:cifar_new} (a) &CIFAR-10  & 5 & 50 & 1 & * & $-$  & CIFAR-10   \\
Figure \ref{fig:cifar_new} (b) &CIFAR-10  & 5 & 50 & 1 & $-$ & * & CIFAR-10   \\
Figure \ref{fig:cifar_new} (c) &CIFAR-10  & 5 & * & 1 & 0.02 & $-$ & CIFAR-10   \\
Figure \ref{fig:cifar_new} (d) &CIFAR-10  & 20 & 5 & * & 0.02 & $-$ & CIFAR-10   \\
 Figure \ref{fig:mae} (c) &CIFAR-10  & 5 & 5 & 1 & * & $-$ & CIFAR-10   \\
Figure \ref{fig:mae} (d) &CIFAR-10  & 5 & * & 1 & 0.02 & $-$ & CIFAR-10   \\
Table \ref{tab:CIFAR10tostl10}  & CIFAR-10 & 5  &5 & 1  & $-$ & $-$  & STL-10   \\

\hline
\end{tabular}}
 
 \caption{\small {\bf Table of experiment setups}. Note that the full experiment pipeline has two steps: pretrain and evaluate. There are two datasets, the pretrain dataset, and the evaluation dataset. For detection tasks on MS-COCO and Amazon, the pretrained encoder is also updating through the transfer learning procedure and for other datasets, only a single linear layer is trained with the pretrained encoder being frozen. $*$ denotes the control variable in each experiment and $-$ denotes the variables that are not used. \vspace{-4mm} }
\label{tab:experiment_setup}
\end{table*}

\section{Additional Experiments} 
\label{appendix:additional_exp}

 \subsection{Additional CIFAR-10 experiments }
We present the experiment details of Dec-SSL  on CIFAR-10, similar to those in Section \S\ref{sec:ssl_vs_sl_constraints} on ImageNet with implementation details in the section \S \ref{appendix:impl} Note that we also include the baseline \textbf{Dec-SL} where the algorithm directly runs  \texttt{FedAvg} on the downstream classification tasks, without explicitly learning a representation. It is known \citep{hsieh2020non} that the non-IIDness is particularly challenging to deal  with in decentralized supervised learning, and we also confirm it in our experiments.

\begin{figure*}[!tb]
\centering 
\includegraphics[width=1\linewidth ]{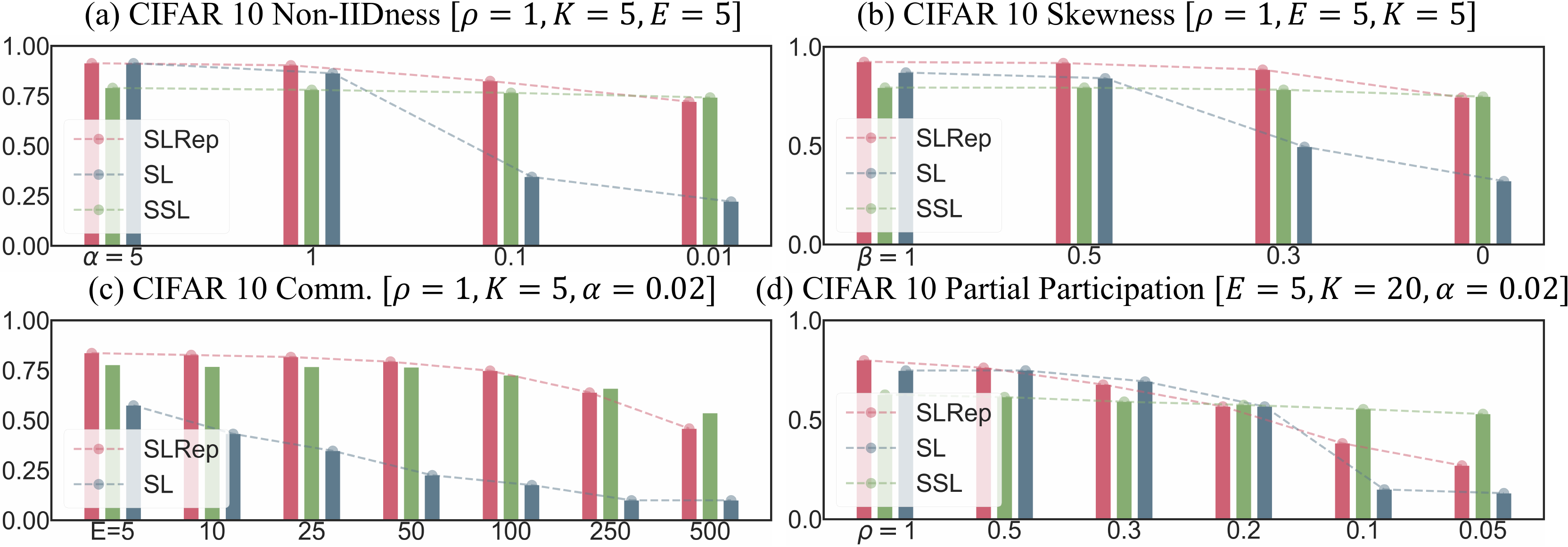} 
  \caption{\textbf{CIFAR-10 Experiments.} SSL is more robust to non-IIDness, communication efficiency, and participation ratios on CIFAR-10 Dataset.}
  \label{fig:cifar_new}
\end{figure*}

\textbf{Decentralized SSL with non-IID data.} On Figure~\ref{fig:cifar_new} (a), we show that Dec-SSL is more robust than Dec-SLRep when we apply Dirichlet label shift to create non-IIDness at different levels. We also observe that  Dec-SLRep outperforms Dec-SL in this decentralized setting. This observation on the CIFAR-10 dataset is consistent with the ImageNet-100 dataset in Section \S\ref{fig:ImageNet}.

\textbf{Decentralized SSL can have better communication efficiency under non-IIDness.}  We use $\alpha=0.02$ in this experiment. Under two different notions of non-IIDness, Dec-SSL is much more robust to communication efficiency compared to Dec-SLRep and Dec-SL.  While the idea of averaging weights after multiple steps sounds challenging, it is surprising to see  how robust Dec-SSL is with respect to the communication frequencies $E$ in Figure~\ref{fig:cifar_new} (b,c). Similar to the ImageNet experiments, Dec-SLRep is less robust to the communication frequencies, and Dec-SL is more brittle to less communication. For CIFAR-10 experiments, each epoch has around 50 iterations.

\textbf{Decentralized SSL allows less participation under non-IIDness.} We use $\alpha=0.02$ in this experiment and fix the total number of epochs to be $500$. We use $K=20$ data sources in this experiment and want to measure the convergence of decentralized algorithms with respect to the participation of data sources at each round. In  Figure~\ref{fig:cifar_new} (d), we show that with non-IID data, SSL is much more robust to less participant each round compared to Dec-SL.  


\begin{figure*}[!tb]
\centering 
\includegraphics[width=1\linewidth ]{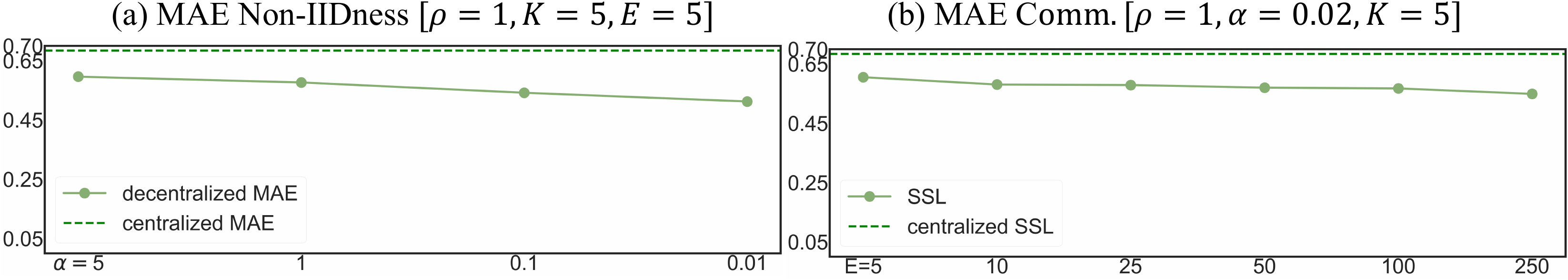} 
  \caption{\textbf{Masked Autoencoder (MAE) Experiments.} We show that the more recent SSL approaches based on masked autoencoder \citep{he2021masked} and vision transformer \citep{dosovitskiy2020image} are also robust to data non-IIDness and communication constraints. This supports that the advantages of Dec-SSL is not restricted to contrastive approaches and convolutional networks.  }
  \label{fig:mae}
\end{figure*}

 {\bf Learning representation  transferable to different data sources.} The idea of transfer learning has been used in the self-supervised learning literature \citep{he2020momentum,pathak2016context} and we apply similar ideas to the decentralized learning setting. In this case, the new data distribution could be treated as {\it a new user/data source}, which we want to perform well and adapt quickly on. We have additional results of linear probing from CIFAR-10 dataset to STL-10 dataset \citep{coates2011analysis} in Table \ref{tab:CIFAR10tostl10}. We found a strong correlation of the downstream classification performance and the transfer learning performance, as they both rely on the representation capacity of the pretrained network.
 
 {\bf Dec-SSL with masked autoencoder.} In this experiment, we run the more recent SSL approach, masked autoencoder, on the CIFAR-10 dataset to investigate its robustness to data non-IIDness as well as communication efficiency.    We use Vit-Tiny \citep{dosovitskiy2020image} with the AdamW \citep{loshchilov2018fixing} optimizer for 1000 epochs with batch size 256. We note that the linear probing performance of MAE is not as good as contrastive learning. However, as shown in Figure \ref{fig:mae}, we still observe a similar stable trend in terms of the downstream performance, as the non-IIDness and the  number of local updates increase. This indicates  that the advantage of Dec-SSL is not restricted to contrastive approaches and convolutional neural networks.

\subsection{Ablation study on dataset and algorithms}
\begin{figure*}[!tb]
\centering
\includegraphics[width=0.8\linewidth]{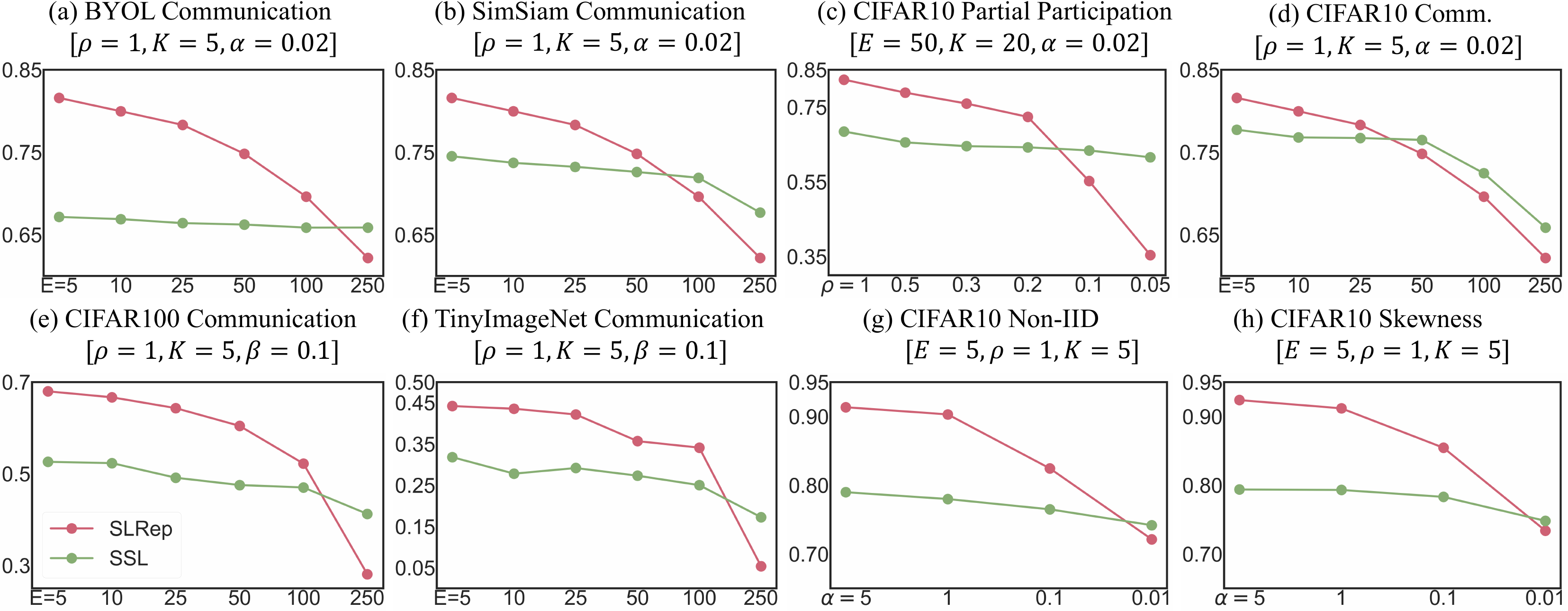} 
  \caption{\small \textbf{SSL method and dataset ablation study.} We conduct ablation study  on  SSL methods  SimSiam and BYOL, as well as on datasets CIFAR-10, CIFAR-100, and TinyImageNet. }
  \label{fig:ssl_ablation}
  
\end{figure*}
\begin{table*}[!t]
\centering
\begin{minipage}[!t]{0.6\linewidth}
{\begin{tabular}{c|ccc}
\hline 
\multicolumn{0}{c|}{CIFAR-10 }      &\multicolumn{3}{c}{STL-10}
  \\ 
 \multicolumn{0}{c|}{ Pretrain}  & \multicolumn{0}{c}{ 100$\%$} & \multicolumn{0}{c}{ 10$\%$} &
  \multicolumn{0}{c}{ 1$\%$} \\ 
  \hline
  
  \color{Gray}no pretrain  & \color{Gray}0.25 & \color{Gray}0.24 & \color{Gray}0.13  \\
Dec-SSL IID  & \textbf{0.65} & \textbf{0.61} & \textbf{0.48} \\
Dec-SLRep IID  & \textbf{0.65} & 0.60 & 0.47 \\
\hline
Dec-SSL Non-IID & 0.60 & 0.54 & 0.36 \\
Dec-SLRep Non-IID & 0.31 & 0.30  & 0.25 \\
Dec-SSL Non-IID Less Comm. & 0.33 & 0.28 & 0.17  \\
\hline
\end{tabular}}
\end{minipage}
\caption{\small Linear Probing from CIFAR-10 to STL10. We observe that pretraining on  non-IID data can negatively affect the performance of transfer learning with different amounts of data. The learned representation from Dec-SSL can improve both the downstream tasks on the same dataset and help transfer to a new dataset.  }
\label{tab:CIFAR10tostl10}
\end{table*}

\label{sec:ablation}
{\textbf{Ablation on SSL algorithms.} We ablate on the learning algorithms in the Dec-SSL setting. We  experiment with  SSL methods SimSiam \citep{chen2021exploring} and BYOL \citep{grill2020bootstrap} in addition to SimCLR to learn representations. From Figure~\ref{fig:ssl_ablation} (a,b), we have consistent observations on the robustness to data non-IIDness, and the stable performance when reducing the communication frequency. These observations confirm that the SSL objectives are in general leading to relatively uniform  features, and are less vulnerable to data heterogeneity with communication constraints.

\textbf{Ablation on dataset.} Furthermore, we ablate decentralized learning on standard datasets such as  CIFAR-100 and Tiny-ImageNet \citep{le2015tiny} and observe that Dec-SSL outperforms Dec-SLRep with communication constraints and non-IIDness (Figure~\ref{fig:ssl_ablation} (e,f)). For CIFAR-10, we also found similar robustness to non-IIDness and skewness (Figure~\ref{fig:ssl_ablation} (g,h)) as well as partial participation and communication constraints  (Figure~\ref{fig:ssl_ablation} (c,d)). On Table \ref{tab:cifar100tocifar10} Bottom, we show that the learned representations from Dec-SSL on one data source (CIFAR-100) can transfer to  other data sources (CIFAR-10)}. Additional ablation study can be found in \S \ref{appendix:additional_exp}.
\subsection{Feature visualization and distance}

\begin{figure*}[!tb] 
\centering 
\includegraphics[width=\linewidth ]{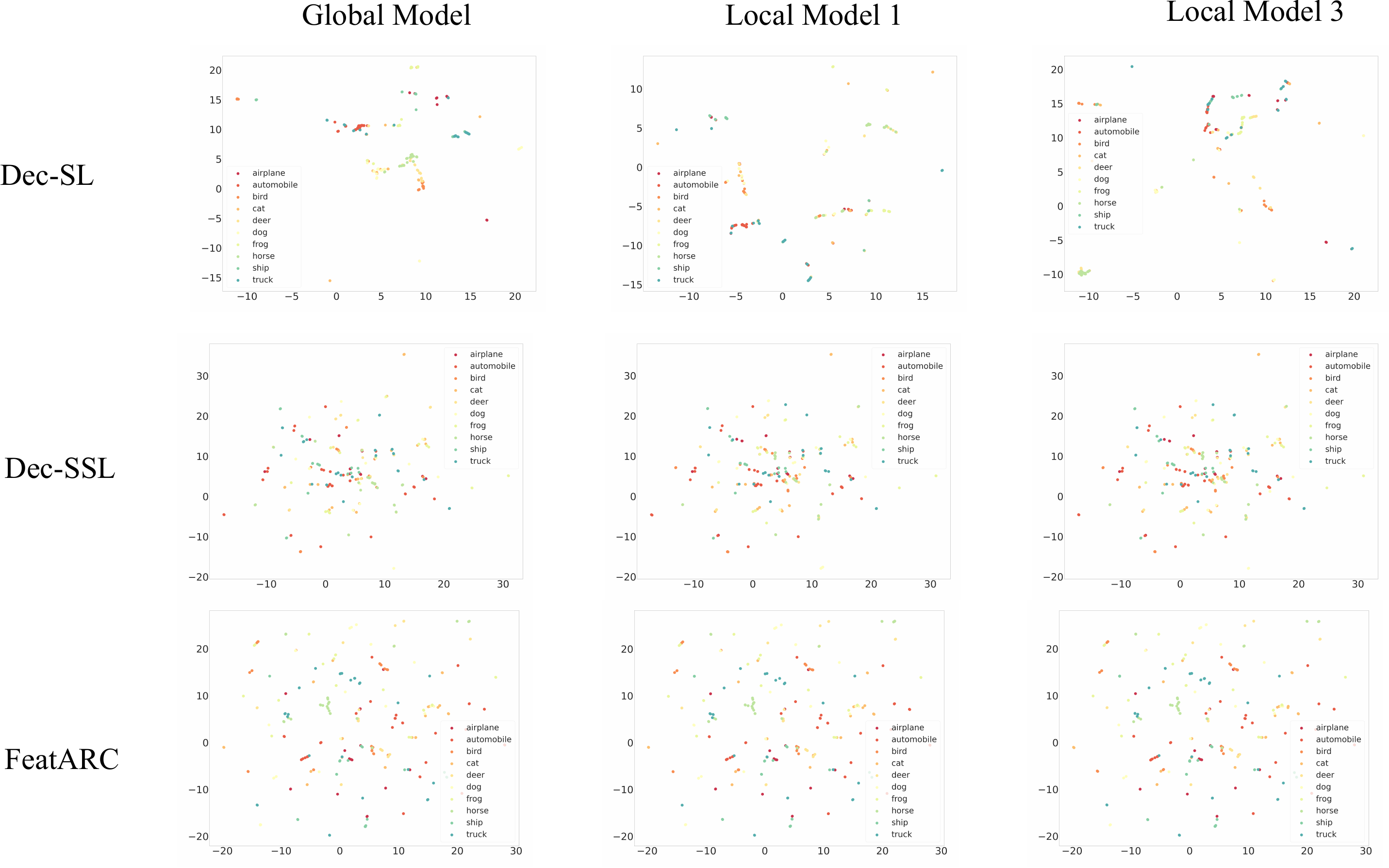} 
  \caption{{\small  \textbf{Visualization of feature space and distance.} {Visualization of the feature space of the local and global models learned from Dec-SSL and Dec-SLRep.}}}

  \label{fig:feat_dist}
\end{figure*}

In Figure~\ref{fig:feat_dist}, we show that the feature generated by three different models: global model $w_g$, local model $w_1$ on data source 1, and local model $w_3$ on data source 3, for both Dec-SSL and Dec-SLRep. We use $5$  local data sources with Dirichlet parameter $\alpha=0.1$ on CIFAR-10 with $500$ rounds with $E=50$ epochs for this experiment. At the final communication round, we have local models $w_1,...,w_5$, and we average to be the global model $w_g$. For Dec-SSL and Dec-SLRep respectively, we first concatenate the features of the three local datasets to plot these three feature sets on the same space. We then use principal component analysis (PCA) to project these features  in $512$ dimensions to $20$ dimensions and use UMap (\cite{mcinnes2018umap}) to visualize these features in $2$ dimension.  

We observe that in Dec-SSL, the features  learned by the local data source are closer to the global model and the features between the local models  are also surprisingly similar to each other; On the other hand, in Dec-SLRep, each data source is learning a drastically misaligned feature space (which can be seen as a visualization of the model itself as well), which matches our theoretical insights in Sections  \ref{sec:theroy} and \ref{sec:theory_2}. 

We also compute the summation of the $\ell_2$-norm difference for each layer of the network weights, denoted as $d_w({\cdot, \cdot})$ as the surrogate for model drift. For Dec-SSL, the weight difference between the global model $w_g$ and local model $w_1$ is $d_w(w_g,w_1)=17.16$, and the weight difference between  $w_1$ and $w_3$ is  $d_w(w_1,w_3)=20.27$, where $w_1$ and $w_3$ correspond to the weights of data source $1$ and $3$, respectively. For Dec-SL or Dec-SLRep, the weight difference is much larger: the weight difference between the global model and the local model $1$ is $d_w(w_g,w_1)=178.92$, and the weight difference between  $w_1$ and $w_3$ is $d_w(w_1,w_3)=202.69$, which is of order larger. For {\tt FeatARC}, the feature  spaces also look aligned and the weight difference (local model 1 and 3 are clustered to global model 1) are $d_w(w_g,w_1)=17.24$ and $d_w(w_1,w_3)=19.71$.

\subsection{$X$-distribution shift among local data sources}


Specifically, in addition to the label and feature distribution shifts in Section \ref{sec:dec_ssl_robust}, in this section, we show that under very non-IID features of the local datasets, the performance of Dec-SSL is {\bf still robust} and {\bf stable}. 

\begin{figure*}[!hb]
\centering 
\includegraphics[width=\linewidth ]{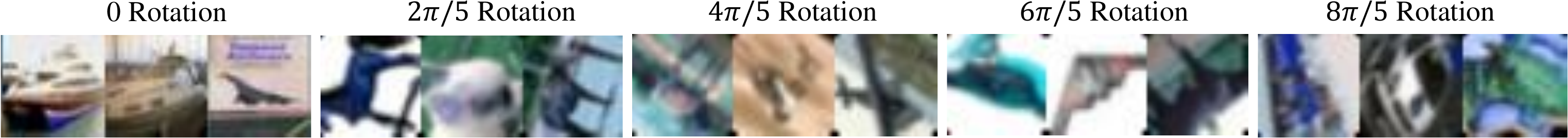} 
  \caption{
  {\small
  \textbf{Visualization of the input distribution shift for CIFAR-10, created with rotation augmentation.} We apply $5$ different rotation augmentation to create the non-IID  data sources \citep{ghosh2020efficient}, which, as the subfigures illustrate, have very different characteristics.} 
  }
  \label{fig:rotation_shift}
\end{figure*}

\begin{figure*}[!hb] 
\centering 
\includegraphics[width=\linewidth ]{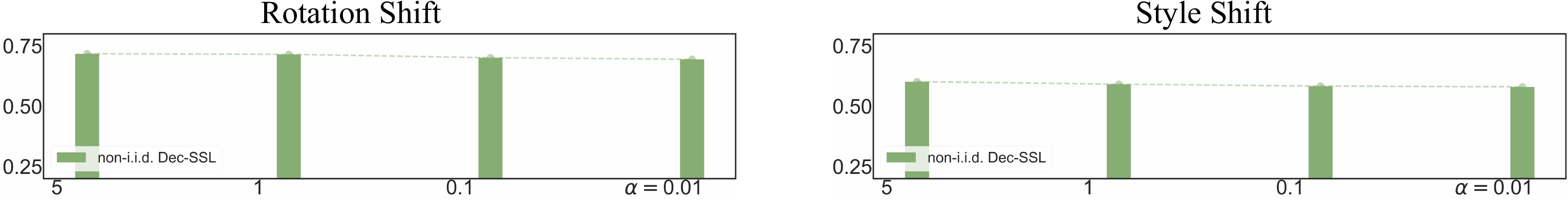} 
  \caption{\small  \textbf{Dec-SSL performance under additional $X$-distribution heterogeneity.} We experiment with two different kinds of input distribution heterogeneity by applying different rotation and stylization distribution heterogeneity to each local data source, as examples of different feature characteristics of the dataset.}
  
  \label{fig:x_shift}
\end{figure*}

In Figure~\ref{fig:x_shift}, we apply {\it rotation} and {\it stylization}  augmentation to the raw dataset to create more different characteristics of the features. For rotation, we would manually alter the dataset input $x$ by rotating it   \citep{ghosh2020efficient}. The source of the heterogeneity $\mathcal{M}$ is therefore the orientation of the images. We split the data into 5 different datasets $D_1,...,D_5$ and apply $0, \frac{2\pi}{5},\frac{4\pi}{5},\frac{6\pi}{5},\frac{8\pi}{5}$ radians of  
rotations to the images in each local data source (see Figure~\ref{fig:rotation_shift} as an  illustration). For stylization \citet{geirhos2018}, we similarly apply $5$ different stylizations to the data in each of the local datasets. As shown in Figure \ref{fig:x_shift}, the same robustness to non-IIDness even when the local dataset has very different features  as above. 

\subsection{Fully decentralized case and  different network topology}\label{sec:full_dec_append}

In this section, we conduct experiments on the {\bf fully decentralized} learning, where the local data sources are only allowed to communicate with their neighbors over a peer-to-peer network, without a centralized server. In short, the observations we had regarding Dec-SSL in the setting with a centralized server still hold, even under several different network topologies. 

\begin{figure*}[!hb] 
\centering 
\includegraphics[width=1\linewidth ]{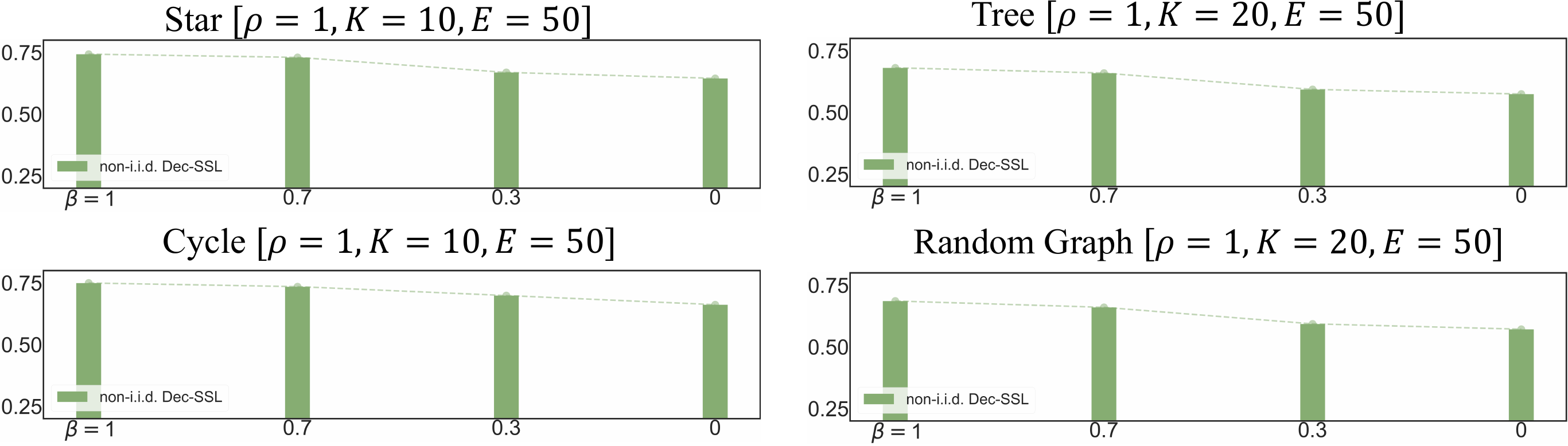} 
  \caption{{\small \textbf{Performance with different topology of the communication networks.} We experiment with four different kinds of network topology in the  decentralized learning setting, and observe a similar behavior of  robustness to the data non-IIDness for Dec-SSL. }}
  \label{fig:topology}
\end{figure*}

In particular, we show that under several different network topologies of the communication networks that connect the  local data sources, the performance of Dec-SSL is stable to the Non-IIDness of the data. In Figure~\ref{fig:topology}, we compare the results with the ``star topology''  (the ``federated learning'' setting with a centralized server and multiple local data sources), the ``cycle topology'', the ``binary tree topology'' with $K=10$ and $K=20$ agents and full participation with $E=50$ epochs, and a random graph with edge probability $0.7$, i.e. there is $0.7$ probability for one edge to appear between two nodes.

These generalization results further  validate the main argument in our main paper, and also align with our theoretical insight provided in Section \ref{sec:dec_ssl_robust}, which came from the benign properties of the {\it solution} to the Dec-SSL {\it objective}, instead  of the {\it specific algorithm} (averaging the iterates via a star or other network topologies) that achieves the solution. These results demonstrate that it is indeed promising to incorporate self-supervision in decentralized learning, even in this peer-to-peer communication case. We hope to further generalize  the results to more complicated ``fully decentralized'' setting in later versions of the paper. 

\subsection{Extremely heterogeneous case  for decentralized learning}

\begin{figure*}[!tb] 
\centering 
\includegraphics[width=1\linewidth ]{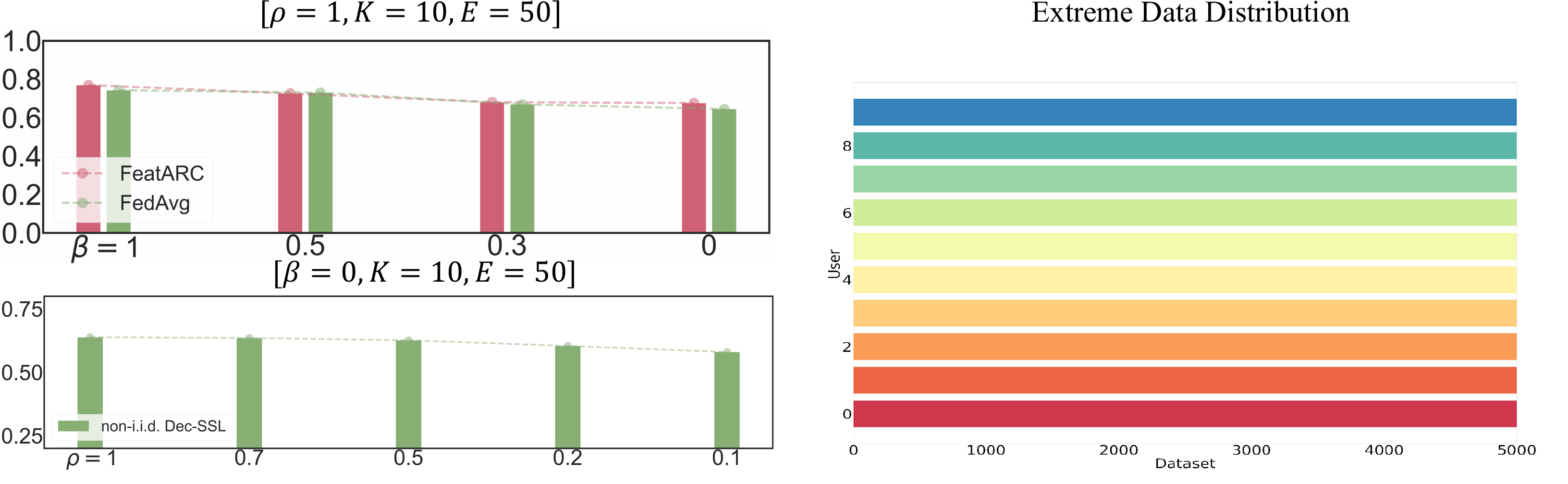} 
  \caption{{\small \textbf{Dec-SSL performance in an extreme case.} In the extremely non-IID case (CIFAR-10) where each local data source owns {\bf only one class} (Right), Dec-SSL still has a robust performance (Left).}}
  
  \label{fig:extre_case}
\end{figure*}

In Figure~\ref{fig:extre_case}, we show that even in the extremely heterogeneous case where each local source only owns one class, the Dec-SSL framework is still robust to the non-IIDness of the data. This is in stark contrast, to the decentralized supervised learning case, which might face an even degenerate classification problem at each local data source with only one class, and its performance is known to degrade. This is also consistent with our theoretical insights in Section \ref{sec:theroy} and Section \ref{sec:theory_2}, and the key reason for the success of Dec-SSL is that, although in terms of ``class'', each local data source only contains a ``unique'' one, but in terms of the information of ``features'' that may be used for the jointly classifying all the  classes, the local data can be rich. In this case, the Dec-SSL objective would not be {\it biased} by the highly heterogeneous  class labels at each local dataset, while the Dec-SL objective could be very much biased by it. See our Figure \ref{fig:theory_plot} for the intuition in a simplified setting.  

\subsection{The effects of data amounts for decentralized learning}
\begin{figure*}[!tb] 
\centering 
\includegraphics[width=0.7\linewidth ]{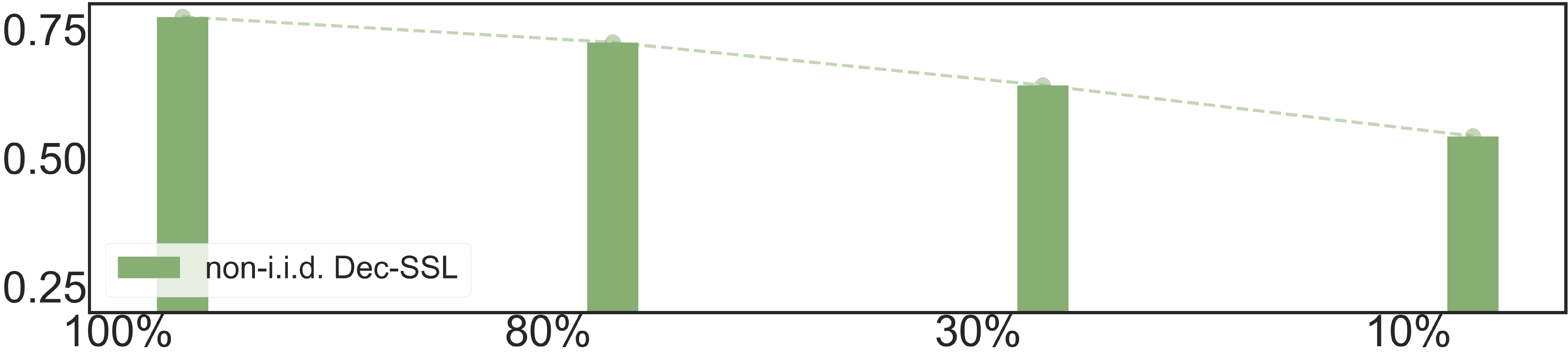} 
  \caption{{\small  \textbf{Relationship of performance  and data amounts.}  {The performance of the Dec-SSL depends on the amount of data for each local data source.}}}
  
  \label{fig:data_amount}
\end{figure*}


We show that the generalization performance depends on the {\it sample size} of the local dataset, which is a motivation for joining federation in  training and is covered in the theoretical formulation. Specifically, In Figure~\ref{fig:data_amount}, we gradually change the data size of the each local data source from $10\%$ to  $100\%$, and observe that the performance (representation power of Dec-SSL) decreases. However, it still maintains $50\%$ accuracy even when each client only owns $10\%$ of the data (in total $50\%$).

\subsection{The effects of data source number for decentralized learning}

\begin{figure*}[!tb] 
\centering  
\includegraphics[width=\linewidth ]{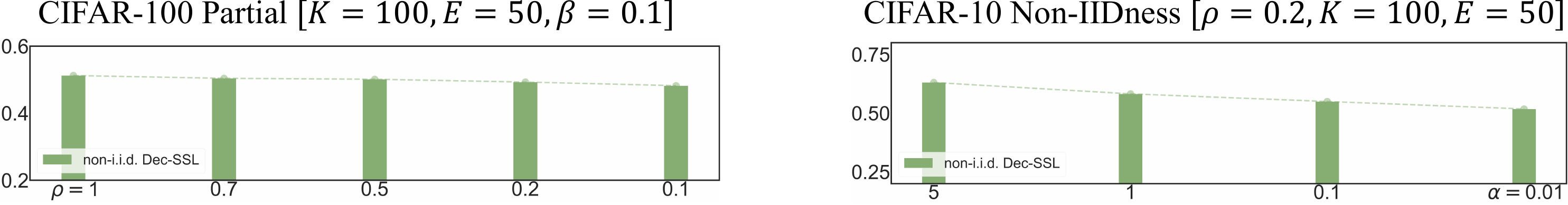} 
  \caption{{ \small \textbf{Dec-SSL performance with more local data sources.} The performance of more data sources participating in the Dec-SSL framework. We here consider a much larger number of local data sources, $K=100$, compared to the $5$ and $20$ used before in the main paper. }}
 
  \label{fig:more_client}
\end{figure*}

We illustrate the results for the experiment with {\it $100$} local data sources  in Figure~\ref{fig:more_client}. We observe that the training performance is still robust to the data non-IIDness on CIFAR-100 (with only 20\% of the participants) and to different levels of partial participation rates on CIFAR-10.  

\subsection{Comparison of FeatARC with other algorithms}

\begin{figure*}[!tb] 
\centering 
\includegraphics[width=1\linewidth ]{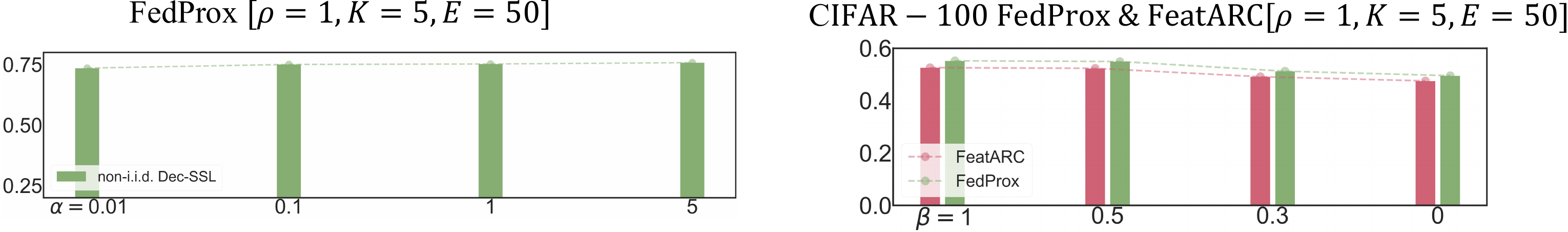} 
  \caption{{\small \textbf{Dec-SSL performance when using other federated learning algorithms.} The performance of Dec-SSL when FedProx  \citep{li2020federated} is used in replace of FedAvg, and its comparison with {\tt FeatARC}.}}
  \label{fig:fedprox}
\end{figure*}

In this section,  we compare our algorithm with Dec-SSL algorithms when combined with other federated learning algorithms, including  \cite{li2020federated} (FedProx) and \cite{li2020fedbn} (FedBN). We note that FedBN  is the underlying implementation for FedAvg in our work, since we simply {\bf did  not} average the batch norm layer during the communication. In Figure~\ref{fig:fedprox} (Left), we show that FedProx also exhibits robustness to the non-IIDness of the data, and we showed that our proposed {\tt FeatARC}. algorithm can outperform FedProx on CIFAR-100 dataset in Figure~\ref{fig:fedprox} (Right).

\section{Method and Experiment Details}
\subsection{Implementation details}
\label{appendix:impl}
 

 
 

 In representation learning, we aim to \textit{pretrain} a network model on a dataset with some pretext tasks and  \textit{transfer} the weights to another problem, potentially a new dataset and a specific downstream task. The most widely and practically used representation is the pretrained weights  supervised learning on ImageNet \citep{he2016deep}, as an initialization for finetuning or training on downstream tasks such as classification and detection. Recently, self-supervised representation learning has attracted 
increasing attention.  It is common to study the performance and behavior of representations through evaluating on downstream tasks. We follow the same setup and try to understand the visual representation learning under the decentralized learning setting. For reference, Table \ref{tab:experiment_setup} shows a list of datasets used for different experiments in  the paper.
 
Unless otherwise noted, we use ResNet18  \citep{he2016deep} throughout the experiments  and train for 500 epochs with the  Adam optimizers\citep{kingma2014adam}, learning rate 0.001, and  batch size 256. We use SimCLR \citep{chen2020simple} as the default SSL algorithm due to its simplicity. For masked autoencoder \citep{he2021masked} experiment on CIFAR-10, we use Vit-Tiny \citep{dosovitskiy2020image} with AdamW optimizer for 1000 epochs with batch size 256. We note that the linear probing performance of MAE is not as good as contrastive learning.   Note that in all experiments, the unit for local update number is epoch instead of iterations (e.g.,  $E=5$  means each local data source would update 5 epochs, about 200 iterations, before averaging). Note that each epoch on CIFAR-10 for $K=5$ data sources is $\delta=50$ iterations and we fix the number of total epochs for all experiments. For ImageNet experiment, we use a learning rate of $0.005$ with $E=200/\rho$ epochs, where $\rho$ is the participation ratio of data sources. For SimSiam and BYOL, we use a learning rate of $0.03$ with the  SGD optimizer. We consider the standard classification benchmark dataset such as CIFAR-10, CIFAR-100 \citep{krizhevsky2009cifar}, ImageNet \citep{krizhevsky2012imagenet}, TinyImageNet  \citep{le2015tiny}, STL-10  \citep{coates2011analysis} and detection dataset such as COCO \citep{lin2014microsoft} and a real-world package detection dataset that comes from Amazon. We only use a subset of the Amazon dataset which has around $80000$ RGB images with contour labels predicted by the Amazon systems. We use the SimCLR  image augmentation for all view augmentation without Gaussian blurring on CIFAR and the standard version on ImageNet. The temperature for SimCLR   is fixed to be $0.5$. For classification tasks,  {to evaluate the learned representation,} we initialize a linear classifier after the feature encoder and train it until convergence on the centralized training set, and then evaluate it on the centralized test set.

For finetuning detectron \citep{Detectron2018} on COCO and Amazon datasets, we use the default schedule with 90000 iterations and the FPN backbone, batch size 16, and learning rate $0.02$. 
We use a centralized dataset whose distribution is the union of all local data sources. For Amazon experiments, recall that each session is considered as a local data source,  and we run pretrain with Dec-SSL with each session trained individually and then communicate. For the evaluation phase on Amazon package detection / segmentation tasks, we train on a subset of $10000$ images of the unlabeled data for $20000$ iterations to show the benefits of representation learning. For this segmentation task, we use the outputs of the Amazon systems as the ``ground-truth'', but we note that they can be inaccurate.  For detection and segmentation tasks, the training and evaluation setups follow those in the  Detectron \citep{Detectron2018}  pipeline, with only the initialization weight being  replaced.

For both  {\tt  FedAvg} \citep{mcmahan2017communication} and {\tt  FeatARC}, we use 5 data sources $(K=5)$ with evenly split number of data per data source. Each round we use full participation $(\rho=1)$ with 5 local update epochs $(E=5)$.  We use step scheduler to gradually decay the learning rates and reset all local optimizer states for each round in the CIFAR-10 experiments, and do not average the BatchNorm layers (FedBN) \citep{li2021fedbn}. In \texttt{FeatARC}, we find 2 clusters to be sufficient to achieve good performance and also use hyperparameter $\lambda=1$. During evaluation, we test on each local dataset using the corresponding cluster model, and average the best performance as the classification accuracy.  All experiments run on one V100 GPU and finish within a day. We use a customized ResNet  to process the CIFAR image, and these experiments take much less resource and time. Note that although we typically compare Dec-SLRep and Dec-SSL on the same dataset, in practice the unlabeled dataset has much larger diversity and quantity. 



\subsection{Data heterogeneity creation details}
\label{appendix:data_hetereneity}

\begin{figure*}[!tb]
\centering 
\includegraphics[width=1\linewidth ]{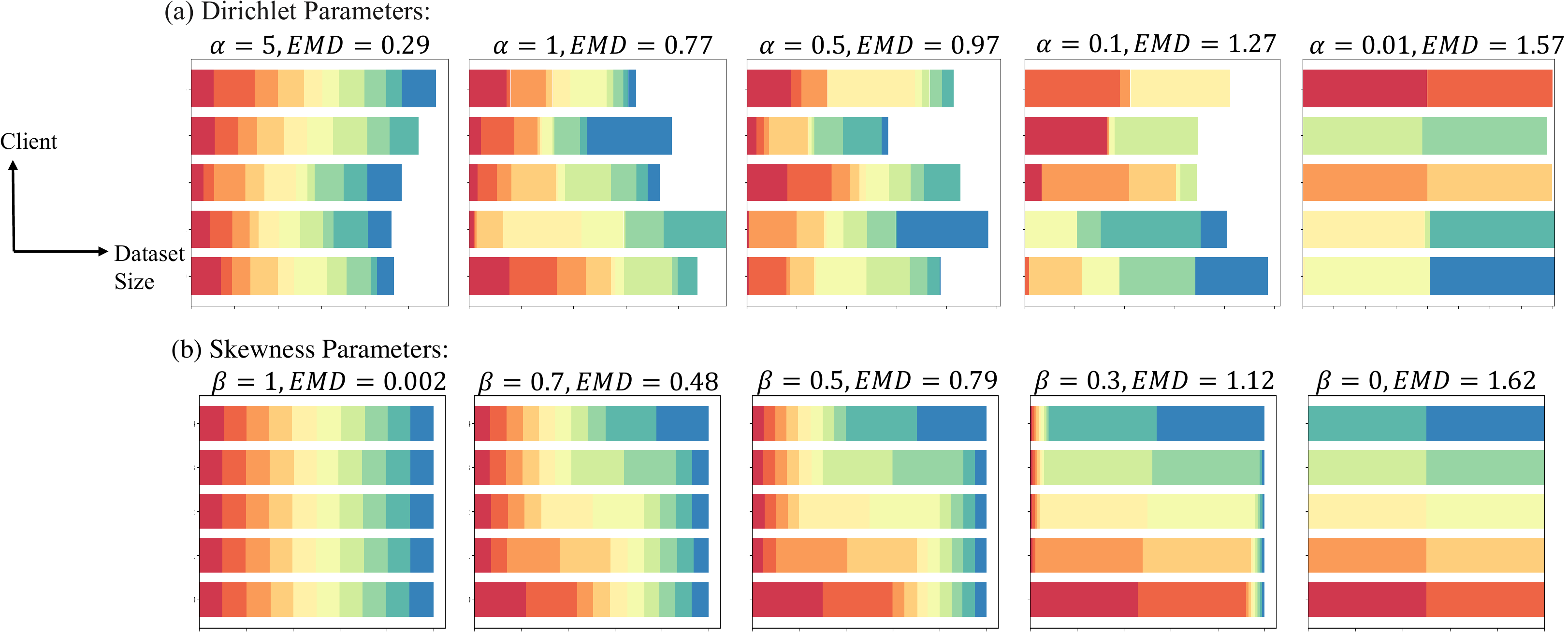} 
  \caption{
  \small
  \textbf{Visualization of the label distribution shift for CIFAR-10.}
  Each horizontal bar represents the data for one data source where one color indicates one class in CIFAR-10 (10 classes in total), and the vertical axis represents different data sources. To study the effect of non-IIDness on decentralized learning, we use Dirichlet process (with parameter $\alpha$) and a skewness ratio (with parameter $\beta$) to split the data. We can observe from left to right that the data becomes more and more non-IID as we adjust the parameters. {We also compute the average earth-mover's  distance from the local dataset to the global dataset to quantify the distribution shift.}
  }
  \label{fig:y_distribution_vis}
\end{figure*}

In this section, we discuss how we construct the non-IIDness of datasets on CIFAR-10. The same procedure applies to other datasets used in the paper. Assume that we split the dataset into $N$ partitions (Note that it is different from the number of data sources $K$), and these  $D_1,..,D_N$ are based on some sources of the heterogeneity. Once we have these $N$ partitions, we use two different ways to create the data non-IIDness across data sources.  The first method is to use a Dirichlet distribution to split $D_1,...,D_N$ \citep{yurochkin2019bayesian}. As a multivariate generalization of the Beta distribution, Dirichlet distribution generates sample $p_k\sim Dir_{N}(\alpha)$ and  assigns  a portion $p_{k,j}$ of the class $k$ to data source $j$. Note that $\alpha$ represents a concentration parameter. When $\alpha$ increases to the limit of $\infty$, the distribution becomes more and more IID (each data source  has roughly a uniform distribution). Empirically for CIFAR-10 with 50000 data points and 10 classes (Figure~\ref{fig:y_distribution_vis}), $\alpha=5$ implies a reasonably uniform distribution over $10$ classes and $\alpha=0.01$ implies an non-IID case each data source has data from mostly $2$ classes and a small amount comes from other uses.  Another way to create non-IIDness is through skewness partitioning \citep{hsieh2020non}. In this case,  we  separate the entire dataset into $(\beta)$  fraction that would split uniformly to each partition and $(1-\beta)$ fraction that would split in a skewed way. Assume we have $N$ partitions, then each data source would have $\beta$ fraction of its data coming from the IID distribution of the dataset, and $(1-\beta)$ fraction that comes from $\floor{N/K}$ of the partitions exclusively. As we decreases $\beta$ from $1$ to $0$, the dataset becomes more heterogeneous. To see this, observe that $\beta=1$ implies that the data is completely uniform from 10 classes and $\beta=0$ means that each dataset has exclusive data from $\floor{N/K}$ of the partitions (2 classes from CIFAR).  Note that for these two approaches, the non-IIDness level is parametrized by $\alpha, \beta$ and in the experiment, we consider a range of $\alpha \in [0.01,5]$ and  $\beta \in [0,1]$. 

\paragraph{Label distribution shift.} This  source of data heterogeneity  comes from the class labels of the data samples. Since CIFAR-10 has 10 classes, we can separate the whole datasets into  $D_1,...D_{10}$ as each $D_i$ contains only the $5000$ images from one class. For instance, $D_1$ can be all cat images and $D_5$ can be all truck images. On Figure~\ref{fig:y_distribution_vis}, we visualize the created non-IIDness on the $y$ distribution (label) by these two approaches.

\begin{figure*}[!tb]
\centering 
\includegraphics[width=1\linewidth ]{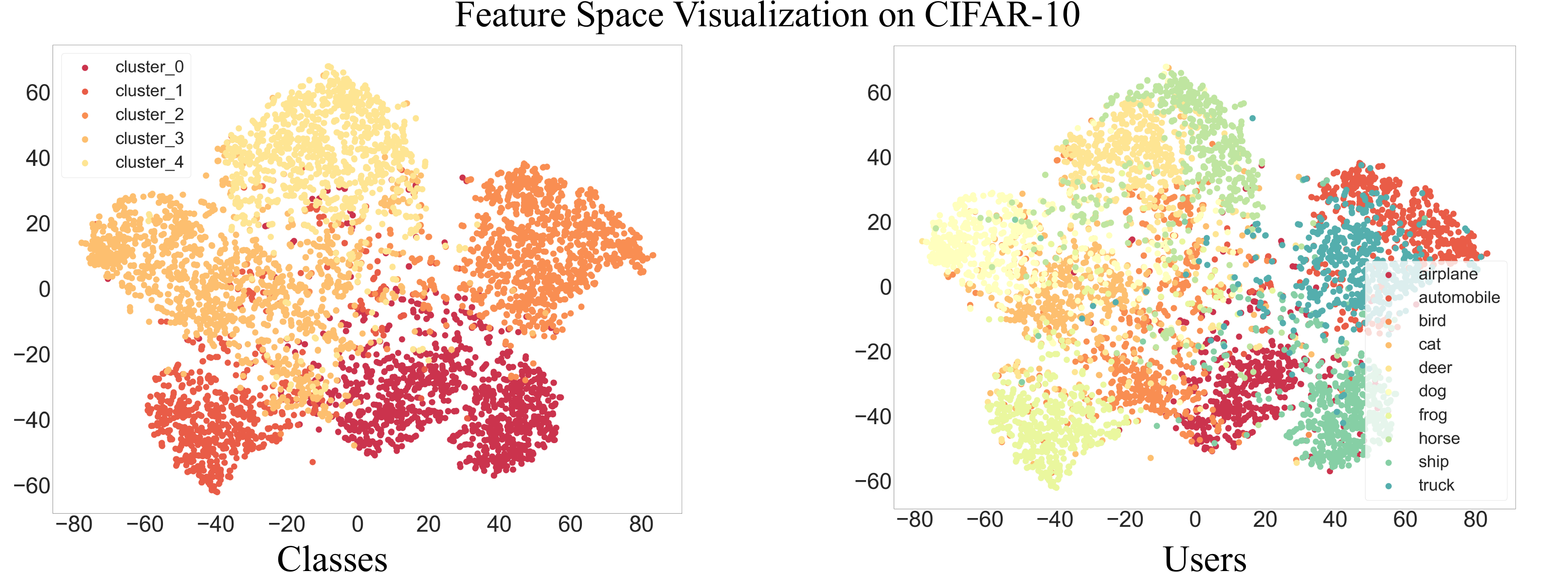} 
  \caption{{\bf Feature distribution shift.} \small Heterogeneity created with a pretrained feature extractor. We use a pretrained network to map images to a feature space and then do clustering to split the dataset.}
  \label{fig:feature_shift}
\end{figure*}

\paragraph{Feature  distribution shift.} This  source of data heterogeneity comes from  the feature space of a pretrained network \citep{zhang2020personalized}. Specifically, we first train a pretrained network on classification task on CIFAR-10 with a ResNet50 and use the $2048$-dimensional  latent vector as a representation of the image feature. After that, we further use Principal Component Analysis (PCA) to reduce the dimension to $30$ and do clustering. Treating each feature space cluster as a partition, we create 5 clusters and visualize the cluster ID and the class ID, in Figure~\ref{fig:feature_shift}.

\subsection{Algorithm details}
  
 \begin{algorithm}[!t] 
 \caption{~Feature Alignment Regularization and Clustering (\texttt{\texttt{FeatARC}})}
\label{alg:farc}
\begin{algorithmic}[1]
\STATE{\textbf{Input}: Cluster number $C$, initialization of  cluster and local models  $\{\theta_j\}_{j\in[C]}$ and  $\{\tilde{\theta}_i\}_{i\in[K]}$
\STATE \textbf{Parameters}:~~Number of local updates $E$, number of  total rounds $T$, distance function $\DD$, learning rate $\gamma$, local datasets  $D_1,...,D_K$} 
\FOR{$t= 0,...,T-1$} 
\STATE \underline{Central server}: Broadcast cluster parameters $\{\theta_j\}_{j\in[C]}$; choose $\cM$, a random subset of data sources to participate at round $t$
\FOR{\underline{Data source} $i\in \cM$ in parallel}
\STATE Initialize local and global feature sets ${z}_{i,1},...,{z}_{i,C},\Tilde{z}_i$  \\
\FOR{$j\in[C]$} 
\FOR{Data sample $x_k \in D_i$ }
\STATE{\text{Compute {\bf global feature}:  }~~${z}_{i,j} \gets {z}_{i,j} \cup \{f_{\theta_j}(x_k)\}$} 
 \STATE {Compute {\bf local feature}  :}~~$\Tilde{z}_i\gets \Tilde{z}_i \cup \{f_{\Tilde{\theta}_i}(x_k)\}$ 
 \ENDFOR 

 \STATE Compute average feature alignments:  $A_{i,j}=\frac{1}{|{z}_{i,j}|}\sum_{k=1}^{|D_i|}\DD({z}_{i,j,k}, \Tilde{z
 }_{i,k})$
\ENDFOR
 \STATE Estimate cluster identity:~~$I_i\gets \arg\min_{j\in[C]}~A_{i,j}$
\STATE{\text{Update {\bf local model}:~~} $\tilde{\theta_i} \gets\texttt{LocalUpdate-FAR}(E,\gamma,\theta_{I_i})$}  
\STATE Send back $\tilde{\theta_i}$ 
and the one-hot vector $s_i=\{s_{i,j}\}_{j\in[C]}$ with $s_{i,j}=\bm{1}_{\{j=I_i\}}$ 
\ENDFOR
 
  \STATE \underline{Central server}: Update cluster model $\theta_j\gets \frac{\sum_{i\in \cM}s_{i,j}\tilde{\theta_i}}{\sum_{i\in \cM}s_{i,j}}$ for all $j\in[C]$ 
  \ENDFOR
 \end{algorithmic}
 
\end{algorithm}
\begin{algorithm}[!t] 
 \caption{~LocalUpdate with Feature Alignment Regularization \texttt{(LocalUpdate-FAR)}}
 \label{alg:alignment_loss} 
\begin{algorithmic}[1] 
\STATE \textbf{Input}:~~Local iteration number $E$,  step size $\gamma$, model $\theta$  
\STATE \textbf{Parameters}:~~SSL objective $L_{\text{SSL}}$, feature distance metric $\DD$,  random augmentation function \texttt{Aug}, balance parameter $\lambda$,  local  dataset $D$
\STATE Set $\tilde \theta \gets \theta$ as the initialization of the {\bf local model}
\FOR{{$t=0,...,E-1$}}
\STATE{\text{Sample data pair:}~~$x$ from ${D}$ and $x^{-}$ from ${D}$ independently}
\STATE{\text{Compute {\bf global feature}:}~~ $z_g\gets f_\theta(x)$}
\STATE{\text{Augment views}:~~$x^+\gets \texttt{Aug}(x)$}   
\STATE{\text{Compute {\bf local feature}}:~~
$(z_l^+,z_l^-,z_l)\gets (f_{\tilde \theta}(x^{+}),f_{\tilde\theta}(x^{-}),f_{\tilde \theta}(x))$}
\STATE \text{Predict feature}:~~ $(p_l^+,p_l^-,p_l)\gets (g_{\tilde \theta}(x^{+}),g_{\tilde\theta}(x^{-}),g_{\tilde\theta}(x))$   
\STATE \text{Compute loss}:~~ $L(\tilde\theta)\gets L_{\text{SSL}}(p_l^+,p_l^-,z_l^+,z_l^-)+\lambda \cdot (\frac{1}{2}\DD(p_l^+,z_g )+\frac{1}{2}\DD(p_l,z_g ))$   
\STATE{\text{Update {\bf local model}}:~~ $\tilde \theta\gets \tilde \theta-\gamma \nabla L(\tilde \theta)$} 
\ENDFOR
\STATE {\bf Return}:~~$\tilde\theta$
\end{algorithmic}
\end{algorithm}    

We here introduce more details about the algorithms we proposed in \S\ref{sec:algorithm_main}. 
Our new algorithm \texttt{FeatARC} is   summarized in Algorithm~\ref{alg:farc};  The subroutine of feature alignment regularization in the local updates is tabulated in Algorithm \ref{alg:alignment_loss}. 
\texttt{FeatARC} is based on the idea of clustering in decentralized learning  \citep{ghosh2020efficient,mansour2020three}, which alternates identifying the cluster identities for each local data source, and using the assigned cluster to do a \texttt{FedAvg} step that averages local models. In federated learning, clustering-based approach is often used as an interpolation between learning local ($K$) models  and learning (a single) global model, in order to tradeoff the bias and variance in learning from heterogeneous datasets  \citep{mansour2020three}. In the highly non-IID scenarios, classic \texttt{FedAvg} with a single global model often fails to capture the heterogeneity of local data distributions, which motivates the use of multiple models, under the assumption that there is some underlying clustering structure of the data (e.g. according to geographic regions, ethnic groups, etc.).  

Specifically, we denote the sets of cluster models and local models as  $\{\theta_j\}_{j\in[C]}$ and  $\{\tilde{\theta}_i\}_{i\in[K]}$, respectively, where $C$ and $K$ denote the number of cluster models and local data sources respectively. 
At each round, we compute an assignment $I_i$ for each local data source $i$ based on matrix $A\in \mathbb{R}^{K\times C}$ where $A_{i,j}$ denotes the ``closeness'' of data source $i$ to  cluster $j$. This ``closeness'' is defined based on how aligned the features are, measured by $\mathbb{D}(\cdot,\cdot)$, between the local feature $f_{\hat{\theta}_i}(x_k)$ and global feature $f_{{\theta}_j}(x_k)$ for each data point $x_k$ in the local dataset $D_i$ (Line 8 to 12 in Algorithm~\ref{alg:farc}). Now to estimate the cluster identity for data source $i$, we use the $\argmin$ over the cluster of the average of feature distance for all data points in the dataset. The assigned cluster model would be sent to the local data source,   and is locally updated with the subroutine  in  Algorithm \ref{alg:alignment_loss}.  Throughout the paper, the distance metric (or the alignment as its negative) between features used in SSL loss, auxiliary loss, and clustering identification, is all defined based on a cosine distance metric $\DD(z_1,z_2)=-\frac{z_1\cdot z_2}{\norm{z_1}\norm{z_2}}$. We use this distance metric, instead of the SSL loss as the metric, since it has been shown that the SSL loss might not be indicative enough for the performance on downstream tasks \citep{robinson2021can}.

 
In  Algorithm \ref{alg:alignment_loss}, we propose to add the distance of the features from the local model to the features from the global model  as an auxiliary loss in the local SSL training, which can be viewed  as distilling global model to the local model, or as a  trust-region update that restricts the drift of local models. Note that we here refer to the cluster model as the ``global model'' in this local subroutine. 
In particular, in addition to the original  self-supervised learning loss $L_{\text{SSL}}$ that takes in positives $x^{+}$ and potentially negatives $x^{-}$, we add a weighted auxiliary loss. The loss is defined as the cosine distance metric on the prediction output $p$ of the local model on data point positives $x^+$, and the feature output $z$ of the global model on data point $x$ (similar to how SimSiam is implemented). This way, when there are many local updates without explicit communication  among local data sources, the global model features can still {\it regularize} the local ones to be close to the global one. 


Compared to other concurrent/contemporaneous  methods,  \citet{zhuang2022divergence} requires an extra memory bank and a customized update rule for local model,  \citet{makhija2022federated}  requires access to an unlabeled public dataset for all models to measure the distance, and requires the communication of some datasets;  \citet{he2021ssfl}  experiments with multiple methods to do personalization in  decentralized learning, and our clustering-based approach can be viewed as new instance of it. In  \texttt{FeatARC}, the auxiliary regularization loss and the clustering procedure between global and local models are simple to add and are general enough to be compatible with any SSL algorithm in decentralized n our experiment, we sweep over the hyperparameters  and choose the balancing hyperparameter $\lambda=1$ and the number of cluster to be $C=2$.

\section{Real-world Decentralized Unlabeled Data Examples}
\label{appendix:discussion}
 \begin{figure*}[!tb]
\centering 
\includegraphics[width=1\linewidth ]{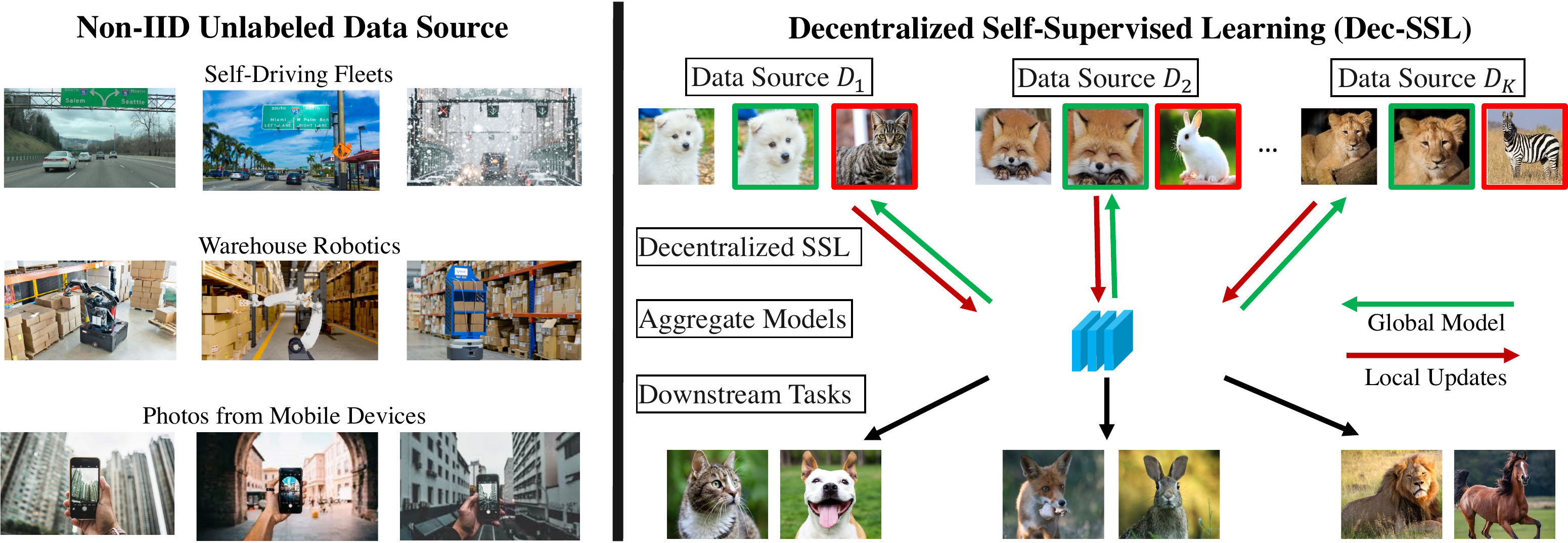}
  \caption{
  \footnotesize
  \textbf{Decentralized self-supervised learning (Dec-SSL).} 
   In the real world, large amounts of unlabeled data are generated and stored in a distributed fashion with high  heterogeneity. 
  In this work, we study decentralized self-supervised learning and apply it to real-world visual representation learning problems.} 
  \label{fig:xz_distribution_vis}
\end{figure*}
 In this section, we enumerate several real-world motivating scenarios where  Dec-SSL with heterogeneous and unlabeled data is relevant and using unlabeled data can play a significant role (Figure~\ref{fig:xz_distribution_vis}). These examples are naturally related to fleets of devices, where model adaptation and data sharing become a central question, and thus require an efficient way to extract information from the decentralized datasets. Note that different from the ``big and diverse data'' motivation for SSL in the centralized setting,  decentralized setting emphasizes that the data come from very distinct data sources,  and the bandwidth in many cases simply cannot afford raw data communications.  
 
\paragraph{Self-driving fleet.} Self-driving cars are naturally deployed around the world with very distinct data distributions. For instance, the traffic rules in Berlin can be very different from the traffic rules in China. The camera observations on a freeway is very different from those on a crowded city road. Despite that data sharing might not be a problem, labeling all masks for images can be a prohibitive tasks and sharing all data can be very inefficient. The data is inherently skewed in terms of quantity $n$, features $x$, as well as labels $y$.   For instance, we can have imbalanced number of classes for an object detector trained to deploy on the freeway that often sees trucks and one trained to deploy on the street that often sees people and cyclists.

\paragraph{Mobile edge devices.} Decentralized supervised learning on the edge devices such as medical diagnosis, object detection, and  sentence completion have been used in the real systems. However, with growing  interests and importance, decision making and interactions with the environment in the wild are more likely to generate unlabeled datasets. With external sensors, one can collect data for agents participating in some tasks such as cooking, doing sports, and working, but we cannot easily provide labels for these settings and these settings can sometimes be privacy sensitive. Take cooking for instance, it can be very difficult to label the masks for all the ingredients and food on the table. Moreover, the data from only each single user might not be enough to learn a generalizable representations through self-supervised learning, thus motivating each user to join a federation, and jointly learn a global model. Thus, it would be very interesting for the community to investigate decentralized self-supervised learning to acquire useful representation from these by nature distributed and diverse data.

\paragraph{Warehouse/Household robots.} A bottleneck in robotics has been the availability of high-quality and large-scale real-world data. As robotic systems are  deployed more and more at scale in both warehouse and households settings, large-scale  datasets are becoming increasingly available. In Section \S\ref{Amazon Data}, we present a detailed example from the actual Amazon warehouse to motivate decentralized self-supervised learning. Similar to self-driving cars, a single robotic work-cell can generate millions of images per year; however, it is impractical to label data at this scale. Moreover, each local data distribution can be narrow and thus the model learned from each local dataset can hardly  generalize. Considering a model trained on data from a warehouse that only sees boxes and then trying to operate this model in a warehouse that sees a variety of package types. To address the overfitting issue, it is useful to learn a {\it common representation}  that can be quickly specialized for each local data source. Our Dec-SSL framework  provides an efficient and robust way to do representation learning. In addition, due to the  communication budget, it is desirable to have longer local updates $E$ and arbitrary participation ratio $\rho$ during the learning process. These methods, taken to full fruition, can enable local systems to efficiently share information and continually improve with significantly fewer labels.  Similarly, a fleet of home robots that are deployed at diverse homes across the world can generate terabytes of raw data that are infeasible to share on cloud databases, due to limits on both the privacy and the network bandwidth. Moreover, the data that is  commonplace for robots at one place can be out-of-distribution for robots at other places, causing challenges on deploying  robots in homes, warehouses, and other human environments. In the next section, we provide more details about the robotics dataset example from Amazon warehouse. 
 \subsection{A real-world  non-IID dataset in robotics} 
\label{Amazon Data}

 \begin{figure*}[!t]
\centering 
\includegraphics[width=1\linewidth ]{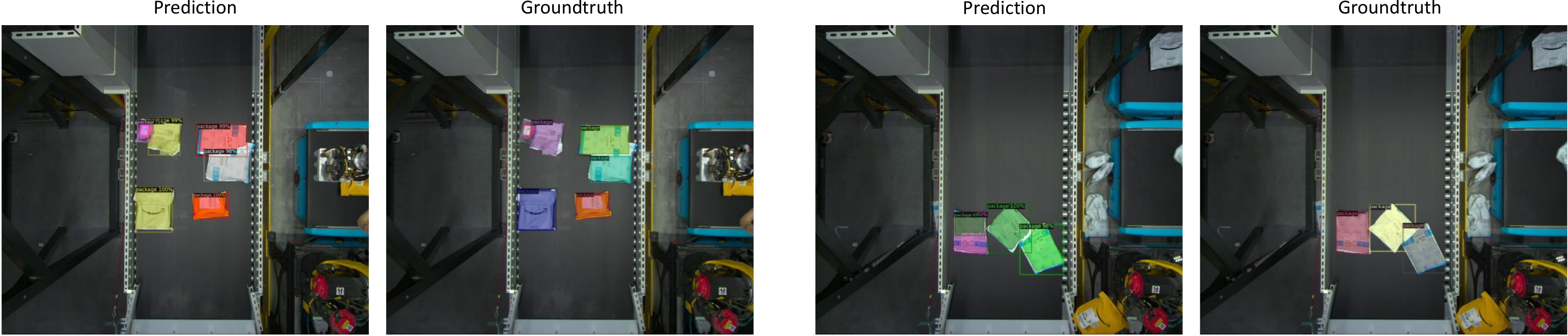}
  \caption{
  \small
  \textbf{Qualitative instance segmentation results on the real world Amazon Data.}
  We applied our decentralized self-supervised learning (Dec-SSL) framework to the real-world data collected in the Amazon warehouses. For each pair of images, the left shows the instance segmentation results of Dec-SSL using a backbone pretrained on the same data, while the right shows the predictions from the Amazon system (used as the ground-truth labels for finetuning). Our method achieves a decent result and outperforms the baseline trained from scratch.  
  }
  \label{fig:detection_label}
\end{figure*}

Robin is a robotic manipulation work-cell at Amazon designed to induct packages into a sortation system. Packages are fed to the robot by means of a conveyor belt and other up-stream material handling equipment. An advanced sensing and perception system on Robin acquires images of the scene, detects and segments packages, and determines what package to pick and how. A custom End of Arm Tool (EoAT) and motion planning and control software robustly execute the pick and place the package on an outbound drive unit. The large-scale deployment of Robin in production provides millions of visual and interaction data. These data are largely unlabeled. As deployments of systems like Robin scale, centrally aggregating data from the entire fleet becomes costly if not infeasible due to bandwidth limits. An additional challenge to continual learning on Robin is that distributions shift at both the individual work-cell level as well as the facility (or site) level, and these shifts present trade-offs in generalization vs. specialization. Said differently, its not clear simply pooling the data is advantageous. The following are some notable ways the Robin dataset is diverse along with several factors that drive this diversity.

\paragraph{Package mix.}
Robin handles many different types of parcels; for example, cardboard boxes, paper bags, poly bags, jiffy mailers, items shipped in their own packaging, etc. A particular facility may see a particular distribution of package types based on its purpose in the network. For example, many sites handle a diversity of package types, weights, and sizes; whereas, other sites may handle predominately only one or two package types or have restrictions based on weight or size.  There are also temporal factors that produce shifts in package type distribution. For example, the introduction of recyclable materials or the use of less packaging material over time.

\paragraph{Package density.}
As is notable in Figure \ref{fig:detection_label}, the density of package presentation varies between facilities and over time. Different sites may have different up-stream material handling systems (e.g. conveyance) that feed Robin packages in different ways. On one extreme, scenes can consist of a single package, and on the other extreme packages are presented in a dense pile with significant overlap and occlusion. During certain times of year (e.g., holiday season) there is increased volume in the network and this can produce denser scenes. 

\paragraph{Hardware configuration.}
Robin work-cells can vary in arm, EoAT, and sensor types throughout the network. Additionally, the size and types of collision geometry in the work-cell area can change at both the work-cell and facility level. These differences mean that even if the input distribution (scenes) are the same, robots may see and move in different ways to accomplish the same task. This further contributes to diversity in both visual and interaction data.  
\begin{figure*}[!tb] 
\centering 
\includegraphics[width=0.9\linewidth ]{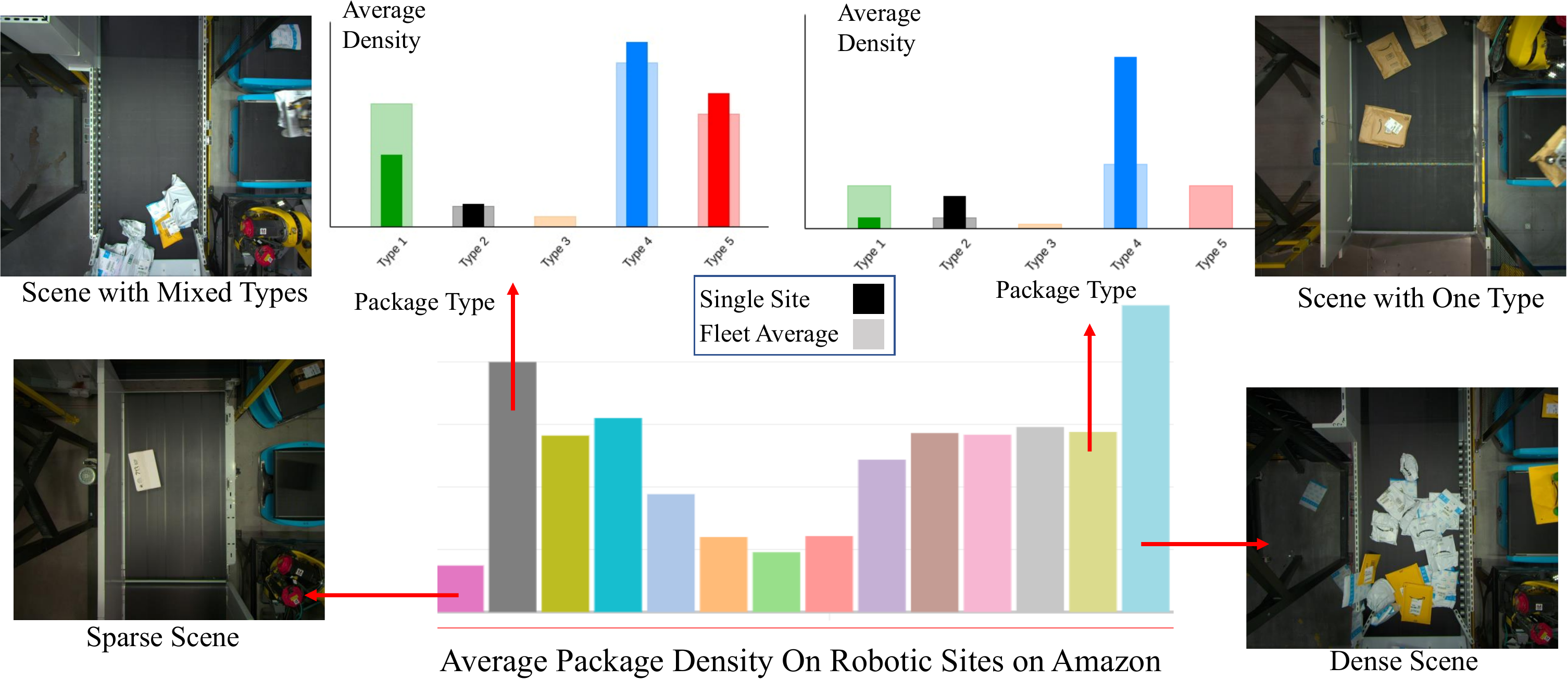} 
  \caption{
  \small
  \textbf{Heterogeneity naturally occurs in real-world  unlabeled data.} In real-world robotic settings such as those in Amazon warehouses, data distribution shifts emerge as a result of differences in location and type of facility, time of year (e.g. holidays), upstream material handling systems, and robotic work-cell configuration (e.g., arm, gripper, and sensor types), etc. The plot shows two axis of non-IIDness across different sites in Amazon: the density of the packages among scenes and the package type distribution within scene compared to the average.}
  \label{fig:amazon_distribution_vis}
\end{figure*}

\section{Theoretical Analysis}
\label{appendix:proof}


\paragraph{Notation.} For any positive integer $k$, we use $[k]$ to denote the set $\{1,2,\cdots,k\}$. We use $\cN(\mu,\Sigma)$ to denote the Gaussian distribution with mean $\mu$ and covariance matrix $\Sigma$. We use $\la x,y  \rangle$ to denote the inner product of two vectors $x,y\in\RR^d$. For two non-negative integers $k,n$, we use $k\bmod n$ to denote the arithmetic remainder of $k$ divided by $n$. For positive integer $d>0$, we use $e_i$ with $i\in[d]$ to denote the $d$ basis vectors in $\RR^d$ Euclidean space. For a real $x\in\RR$, we use $\floor{x}$ and $\ceil{x}$ to denote the floor and ceiling integers of $x$, i.e., $\floor{x}=\max\{k\in\ZZ\given k\leq x\}$ and $\ceil{x}=\min\{k\in\ZZ\given x\leq k\}$.

In this section, we aim to shed some lights on the robustness  of decentralized SSL approaches to data heterogeneity, and their comparison to decentralized supervised learning.  

\subsection{Deferred details and proof  in Section \S\ref{sec:theroy}}\label{sec:ssl_proof}

\textbf{Setup.} Consider a decentralized SSL problem with $K$ data sources.  To model non-IIDness across them, we use a common type of data heterogeneity, i.e., the label heterogeneity (see our discussions in \S\ref{appendix:data_hetereneity}). Indeed, label heterogeneity has been recognized as a fundamental and pervasive problem for decentralized learning, causing  significant performance loss across many applications \citep{hsieh2020non}. This setting also corresponds to one non-IIDness we used in previous subsections (see e.g., \S\ref{sec:CIFAR_noniid}). Similar to the SimSiam approach \citep{chen2020simple}, we first augment $x$, an anchor sample from the dataset to have two positive samples, by sampling $\xi,\xi'\sim\cN(0,I)$ IID from the Gaussian distribution. Consider the linear embedding function $f_w(x)=wx$, where $w\in\RR^{m\times d}$ and $m\geq 2K$. The local SSL objective for data source $k$ is given by
\#\label{equ:local_obj_theory}
\cL_k(w):=
-\hat{\EE}\big[(w(x_{k,i}+\xi_{k,i}))^\top(w(x_{k,i}+\xi_{k,i}'))\big]+\frac{1}{2}\|w^\top w\|_F^2, 
\#
where $\hat{\EE}$ is taken expectation over the dataset $x\sim D_k$, and the randomness of  $\xi_{k,i}$ and $\xi_{k,i}'$.  
Moreover, recall the global objective is given in \eqref{equ:obj_1}. Note that \eqref{equ:local_obj_theory} instantiates the SimSiam loss with the negative inner-product $\la a, b\rangle$ as the distance function $\DD(a,b)$ 
and no feature predictor for simplicity.   We also add a regularization term $\|w^\top w\|_F^2/2$ to improve the mathematical  tractability of the objective (which in practice corresponds to weight decay in the optimizer). The $K$ data sources collaboratively minimize \eqref{equ:obj_1}, and evaluate the learned representation on a $2K$-way classification task. 

\paragraph{Data heterogeneity.} Choose $K=\Theta(d^{1/20})$. The $K$  local datasets are generated as follows. For a fixed data source $k$, the labels are skewed in that data from classes $2k-1$  and $2k$ constitute the majority of the data, while other classes are rare, or even unseen. Specifically, let $e_1,\cdots,e_d$ denote the standard unit basis of $\RR^d$, and let $n_{k,j}$ for $j\in[2K]$ denote the number of data for class $j\in[2K]$ in this  dataset $k$.  
For class $2k-1$, data is generated following $x^{(2k-1)}=e_k- \sum_{i\neq k,i=1}^K q^{(2k-1,i)}\tau e_{i}+\mu  \xi^{(2k-1)}$, where $q^{(2k-1,i)}$ are sampled uniformly from $\{0,1\}$, $\xi^{(2k-1)}\sim\cN(0,I)$, and both $\tau=d^{1/5}$ and $\mu=d^{-1/5}$ are positive hyperparameters.  Similarly, for class $2k$, $x^{(2k)}=-e_k- \sum_{i\neq k,i=1}^K q^{(2k,i)}\tau e_{i}+\mu  \xi^{(2k)}$. The amounts of data from classes $2k-1$ and $2k$ are equal and  both of order $\texttt{poly}(d)$. For classes $2i-1$ with $i\neq k$, $x^{(2i-1)}=e_i+\mu\xi^{(2i-1)}$, and there is no data for classes $2i$ in data source $k$. The amounts of data in classes $2i-1$ with $i\neq k$ are the same and of order sublinear in $d$, i.e., $O(d^{\beta})$ for some $\beta\in(0,1)$, such that $O(K d^{\beta})\leq O(d^{1/5})$. Note that this leads to that $O(K d^{\beta}/(2n_{k,2k}))\leq O(d^{-4/5})\leq O(1)$, and  implies that the sum of the data from all the infrequent    classes $2i-1$ and $2i$ for $i\neq k$ are less than the data in the frequent classes $2k-1$ and $2k$. All $K$ local datasets are assumed to contain the same amount of data, i.e., $|D_1|=|D_2|=\cdots=|D_K|$.  
We visualize the heterogeneous data distribution in Figure \ref{fig:theory_plot}. 


\paragraph{Proof of Theorem \ref{thm:main}:}
\paragraph{For local dataset $k$.} We first analyze the solution to  minimizing the local objective \eqref{equ:local_obj_theory}, using only local dataset $D_k$.  Define  $$X_k:=\hat{\EE}_{x \sim D_k}(xx^\top)=\frac{1}{|D_k|}\sum_{i=1}^{|D_k|}x_{k,i}x_{k,i}^\top$$ to be the empirical data covariance matrix for dataset $k$. Notice that 
{\small
\#\label{equ:EX_k}
&\EE(X_k)\\ 
&=\diag\Big(\underbrace{\tau^2+O(d^{-2/5}),~\tau^2+O(d^{-2/5}),\cdots,~\underbrace{1+O(d^{-2/5})}_{k\text{-th}~\text{term}},\cdots,\tau^2+O(d^{-2/5})}_{K~\text{terms}},~\underbrace{O(d^{-2/5}),\cdots,O(d^{-2/5})}_{d-K~\text{terms}}\Big)\notag\\  
&=\diag\Big(d^{2/5}+O(d^{-2/5}),\cdots,~1+O(d^{-2/5}),~\cdots,~d^{2/5}+O(d^{-2/5}),~O(d^{-2/5}),\cdots,O(d^{-2/5})\Big).\notag
\#}

By matrix concentration bounds, e.g., \citep{vershynin2018high}, we have that with probability at least $1-\frac{1}{2}e^{-d^{1/10}}$, $\|X_k-\EE(X_k)\|\leq O(d^{-2/5})$. By
Weyl's inequality we have that with high probability, 
\#\label{equ:eigenvalue_concentra}
|\lambda_{k,i}-\lambda_i(\EE(X_k))|\leq \|X_k-\EE(X_k)\|_2\leq O(d^{-2/5})
\# 
for all $i\in[d]$, where we denote $\lambda_{k,i}:=\lambda_i(X_k)$ as the $i$-th largest eigenvalue of $X_k$. 

On the other hand, as   $|D_k|\geq \texttt{poly}(d)$,  for any $e_j$ with $j\in[K]\setminus\{k\}$, we have that with probability at least $1-\frac{1}{2}e^{-d^{1/10}}$,  at least $1/3$ (where at least  $2/3$ data come from classes $2k-1$ or $2k$, and $1/2$ of them) satisfy that either 
 $q^{(2k-1,j)}$ or $q^{(2k,j)}$ is $1$, and $\sum_{i=1}^{|D_k|}|\la\xi_{k,i},e_k\rangle\big|/|D_k|\leq O(d^{1/10})$\footnote{Note that we here slightly abuse the notation by denoting the noise in generating the data point $x_{k,i}$ by  $\xi_{k,i}$, which should not be confused with the augmentation noise in the SSL objective \eqref{equ:local_obj_theory}.}  
 (see \citep[Lemma E.1]{liu2021self}). Hence, we have that 
 \small
\#\label{equ:quadratic_e_j}
e_j^\top X_k e_j= \hat{\EE}_{x\sim D_k} \big[(e_j^\top x)^2\big]\geq \big[\hat{\EE}_{x\sim D_k} (e_j^\top x)\big]^2\geq \Big(\frac{1}{3}\tau-\mu \sum_{i=1}^{|D_k|}\frac{1}{|D_k|}\Big|e_j^\top  \xi_{k,i} \Big|\Big)^2=\Omega(\tau^2)=\Omega(d^{2/5}),
\#
\normalsize
with probability at least $1-\frac{1}{2}e^{-d^{1/10}}$, 
where we use the fact that $\mu=d^{-1/5}$. 

Now notice that the local objective in \eqref{equ:local_obj_theory} can be equivalently re-written as 
\#\label{equ:local_rewritten}
\min_{w}~\|X_k-w^\top w\|_F^2,
\# 
which, by  Eckart-Young-Mirsky theorem \citep{eckart1936approximation}, yields that the span of the rows of optimal $w$ (an $m\times d$ matrix) is the span of the eigenvectors of the  first $m$ eigenvalues of $X_k$. Let $\{v_{k,1},\cdots,v_{k,d}\}$ denote the set of $d$ orthonormal eigenvectors of $X_k$, then $X_k=\sum_{i=1}^d\lambda_{k,i} v_{k,i}v_{k,i}^\top$, where recall that $\lambda_{k,i}$ is the $i$-th largest eigenvalue of $X_k$. Hence, by \eqref{equ:quadratic_e_j}, we have
\#\label{equ:apply_quadratic_lb}
\lambda_{k,1}\sum_{i=1}^d(e_j^\top v_{k,i})^2\geq e_j^\top X_k e_j=\sum_{i=1}^d\lambda_{k,i}(e_j^\top v_{k,i})^2\geq \Omega(d^{2/5})
\# 
with high probability. In fact, by   \eqref{equ:eigenvalue_concentra} and \eqref{equ:EX_k}, we can have finer bounds of $e_j^\top X_k e_j$ as 
\#
& e_j^\top X_k e_j= e_j^\top \EE(X_k) e_j+ e_j^\top \big[X_k-\EE(X_k)\big] e_j\notag\\
&\geq d^{2/5}+O(d^{-2/5})-\|X_k-\EE(X_k)\|\geq   d^{2/5}-O(d^{-2/5}),\label{equ:apply_quadratic_lb_2_lb}\\
& e_j^\top X_k e_j= e_j^\top \EE(X_k) e_j+ e_j^\top \big[X_k-\EE(X_k)\big] e_j\notag\\
&\leq d^{2/5}+O(d^{-2/5})+\|X_k-\EE(X_k)\|\leq   d^{2/5}+O(d^{-2/5}),\label{equ:apply_quadratic_lb_2_ub}
\#
where we use the fact that $\|X\|_{\max}\leq \|X\|$ for symmetric $X$. 

Furthermore, by   \eqref{equ:eigenvalue_concentra} and \eqref{equ:EX_k}, we know that
\#\label{equ:apply_quadratic_lb_4}
&d^{2/5}-O(d^{-2/5})\leq \lambda_{1}(\EE(X_k))-O(d^{-2/5})\leq \lambda_{k,1}\notag\\
&\quad\leq \lambda_{1}(\EE(X_k))+O(d^{-2/5})=d^{2/5}+O(d^{-2/5}) 
\#
showing that $\lambda_{k,1}=d^{2/5}\pm O(d^{-2/5})$. Combining \eqref{equ:apply_quadratic_lb_2_lb} and \eqref{equ:apply_quadratic_lb_4}, we obtain that
\#\label{equ:proj_e_j_bnd}
\sum_{i=1}^d(e_j^\top v_{k,i})^2\geq \frac{d^{2/5}-O(d^{-2/5})}{d^{2/5}+O(d^{-2/5})}\geq 1-O(d^{-4/5}),
\#
which completes the proof with $\alpha=4/5$. 


\paragraph{For global dataset.}
Recall the global objective given in \eqref{equ:obj_1}: 
\#\label{equ:global_obj_recall_old}
\min_{w}~~\sum_{k\in[K]}\frac{|D_k|}{|D|}\cL_k(w).
\#
As the local objective in \eqref{equ:local_obj_theory} can be equivalently re-written as \eqref{equ:local_rewritten}, we can also re-write the global objective as 
\#\label{equ:global_obj_recall}
\min_{w}~~g(w):=\sum_{k\in[K]}\frac{|D_k|}{|D|}\|X_k-w^\top w\|_F^2.
\#
Further, note that the gradient of $g(w)$ in  \eqref{equ:global_obj_recall} at any $w$ is the same as that of the following objective:
\#\label{equ:equiv_new_global_obj}
\tilde g(w):=\bigg\|\underbrace{\sum_{k\in[K]}\frac{|D_k|}{|D|}X_k}_{\bar X}-w^\top w\bigg\|_F^2. 
\#
Thus, these two objectives share the same minimizer. Note that it is the minimizer that we care about (as it determines the feature mapping), and  minimizing \eqref{equ:equiv_new_global_obj} is equivalent to minimizing the SSL objective over the global dataset $D=\bigcup_{k\in[K]}D_k$, with the empirical data covariance matrix 
\#\label{equ:global_covariance}
\bar{X}:=\hat{\EE}_{x \sim D}(xx^\top)=\frac{|D_k|}{|D|}\sum_{k\in[K]}\frac{1}{|D_k|}\sum_{i=1}^{|D_k|}x_{k,i}x_{k,i}^\top=\frac{1}{|D|}\sum_{i=1}^{|D|}x_{i}x_{i}^\top.
\#
Hence, \eqref{equ:global_obj_recall} is equivalent to solving
\#\label{equ:global_obj_reform}
\min_{w}~~\tilde g(w)=\|\bar{X}-w^\top w\|_F^2.
\#
We can now follow the analysis above. First, by \eqref{equ:EX_k} and the linearity of expectation, we have 
{\small
\#\label{equ:EX_bar}
&\EE(\bar X)\\ 
&=\diag\Big(\underbrace{d^{2/5}-\Theta(d^{7/20})+O(d^{-1/20}),\cdots,~d^{2/5}-\Theta(d^{7/20})+O(d^{-1/20})}_{K~\text{terms}},~O(d^{-2/5}),\cdots,O(d^{-2/5})\Big),\notag
\#}
 where we have used the fact that 
$$
\frac{(K-1)\cdot d^{2/5}+1}{K}=(1-\Theta(d^{-1/20}))\cdot d^{2/5}+O(d^{-1/20})=d^{2/5}-\Theta(d^{7/20})+O(d^{-1/20}). 
$$
Then, by similar arguments from \eqref{equ:eigenvalue_concentra}-\eqref{equ:apply_quadratic_lb_2_ub}, we have that for all $j\in[K]$ (without excluding any $k$), 
\#
& e_j^\top \bar X e_j\geq d^{2/5}-\Theta(d^{7/20})+O(d^{-1/20})-O(d^{-2/5}),\label{equ:apply_quadratic_lb_2_lb_barX}\\
& \lambda_{1}(\bar X)\leq \lambda_{1}(\EE(\bar  X))+O(d^{-2/5})=d^{2/5}-\Theta(d^{7/20})+O(d^{-1/20}) \label{equ:apply_quadratic_lb_2_ub_barX}
\#
leading to that
\$
\sum_{i=1}^d(e_j^\top \bar v_{i})^2\geq \frac{d^{2/5}-\Theta(d^{7/20})+O(d^{-1/20})-O(d^{-2/5})}{d^{2/5}-\Theta(d^{7/20})+O(d^{-1/20})}\geq 1-2\cdot O(d^{-4/5}),
\$
for large enough $d$ such that $1-O(d^{-1/20})\geq 1/2$, where $\{\bar v_{1},\cdots,\bar v_{d}\}$ denote the $d$ orthonormal  eigenvectors of $\bar X$. 
This completes the proof. 
\hfil\qed

\subsection{Deferred results and proof in Section  \S\ref{sec:ssl_vs_sl_constraints}}\label{sec:sl_proof}

\paragraph{Setup.} The data  are generated as  in \S\ref{sec:ssl_proof}. For each local dataset $k$, consider a supervised learning algorithm that uses a two-layer linear network $g_{u_k,v_k}(x):=v_k u_kx$   as classifier, where $u_k\in\RR^{m\times d}$ and $v_k\in\RR^{c\times m}$ for some $m\geq c= 2K$ are weight matrices. Note that $u_k x$ can be viewed as the feature learned by this classifier, which can be used in the  downstream tasks. This is exactly the protocol of {\bf Dec-SLRep} on the local objective. Following \cite{liu2021self}, we consider the approach of learning the network with minimal norm $\|(u_k)^\top u_k\|_F^2+\|(v_k)^\top v_k\|_F^2$ subject to the margin constraint that $[g_{u_k,v_k}(x)]_y \geq [g_{u_k,v_k}(x)]_{y'}+1$ for all data $(x,y)$ in the local dataset $k$ with all $y'\neq y$. Note that such a solution can be found in direction via gradient descent using logistic loss \citep{ji2018gradient}. 
Now we are ready to  show the following result, based on the   techniques in \cite{liu2021self}. 




 \begin{proposition}[Representations learned by Dec-SLRep across  heterogeneous data sources] 
 \label{prop:result_sl}
With high probability, the feature matrix  $u_k=[u_{k,1},\cdots,u_{k,m}]^\top\in \mathbb{R}^{ m \times d}$ learned from the local dataset $D_k$ has the following properties:  
$$
\sum_{i=1}^m\langle u_{k,i}, e_{j} \rangle^2\le O(d^{-\frac{1}{10}}),
$$
 for $j\in[K]\setminus \{k\}$; while 
$$
\sum_{i=1}^m\langle u_{k,i}, e_{k} \rangle^2\ge 1-O(d^{-\frac{1}{20}}).
$$ 
In other words, the correlation between the learned features in $w_k$ and $e_j$ is small for all $j\in[K]\setminus \{k\}$, while the correlation between the features and $e_k$ is large. 
\end{proposition}  

The proposition suggests that the learned features for each local dataset $k$ overfit the its skewed   data, and does not learn the feature directions, e.g., other unit vector directions $e_j$ for $j\in[K]$ and $j\neq k$, that might generalize well to the data in  other datasets. The result can be viewed as a multi-class generalization of the first part of Theorem 3.1 in \cite{liu2021self}
. The intuition is also  illustrated in Figure \ref{fig:theory_plot}. This way, the feature space learned from various local datasets differ significantly, in that most of the directions among $\{e_1,\cdots,e_K\}$ are uninformative, while their possibly  informative feature directions are all different. This heterogeneity between local solutions is not in favor of  {\it local updates}, as too many local updates would drift the iterates towards its local optimum, and the iterates would become too far away from each other, hurting the convergence  of classic decentralized learning algorithms    as \texttt{FedAvg}. Hence, compared with the Dec-SSL case and  Theorem \ref{thm:main},   Dec-SLRep can be less  communication-efficient as it does not allow large number of local updates. 


\paragraph{Proof of Proposition \ref{prop:result_sl}.}
Without loss of generality, we show the result for dataset $D_1$, i.e., when $k=1$. The proof follows mostly from the proof of Theorem 3.1 in \cite{liu2021self}, and for conciseness, we only layout the key differences. For convenience, we remove the index $k$ in the notation whenever it is clear from the context. First,  note  that the local SL problem is equivalent to the following one:
\#\label{equ:local_SL_equi}
\min_{w}~~\sum_{i=1}^c\|\tilde w_i\|^2_2~~~~~~{s.t.}~~~~\la \tilde w_y,x\rangle\geq \la \tilde w_{y'},x\rangle +1,~~\forall~ (x,y)\in D_1,~~~y'\in[2K],~y'\neq y,
\#
where $\tilde w=[\tilde w_1,\cdots,\tilde w_c]^\top$. We then establish the following lemma. 

\begin{lemma}[Margin \& norm bounds]\label{lemma:margin_gap}
Given the data generated above. Construct a solution to  \eqref{equ:local_SL_equi} as  $w_1^*=e_1,w_2^*=-e_1$, and for $i\in\{2,\cdots,K\}$,  $w^*_{2i-1}=\frac{1}{\mu d }\sum_{j=1}^{n_{2i-1}}\xi^{(2i-1)}_j$ and $w_{2i}^*=\bm{0}$. Then we have that for large enough $d$, with probability at least $1-e^{-d^{1/10}}$, the margin of $\{w_1^*,\cdots,w_{2K}^*\}$ is at least $1-O(d^{-1/10})$. Moreover, we have $\|w^*_{j}\|_2^2\leq O(d^{-3/10})$ for $j\in[2K]\setminus\{1,2\}$. 
\end{lemma}
\noindent\textbf{Proof sketch.} The proof follows from the proof of Lemma E.2 in \cite{liu2021self}. The argument for the data in classes $1$ and $2$ is similar; the argument for that in classes $2i-1$ for $i\in\{2,\cdots,K\}$ is similar to that for class $3$ in the proof therein. Note that the total number of data in the rare classes here equals that of the rare class $3$ therein, which is $O(d^{1/5})$. So one needs to replace the $n_3$ therein by $O(d^{1/5})/K=O(d^{3/20})$ (recall that $K=\Theta(d^{1/20})$), which is a smaller number that 
validates the arguments in the proof therein, and in fact, makes the norm of $\|w_j^*\|_2$ smaller, i.e., $\|w_j^*\|_2\leq O(d^{-3/20})$.   Also,  note that there is no margin constraints corresponding to  classes $2i$ with $i\in\{2,\cdots,K\}$, as there is no data belong to these classes in this local dataset. Finally, note that for any data $(x,y)$ in the dataset, $x^\top w_{2i}^*=0$, which does not affect the margin between  other classes and $2i$. The remaining of the proof follows from the proof therein. \hfill\qed 



Then, similar to the argument in the proof of Theorem 3.1 in \cite{liu2021self} (supervised learning part), one can show that by normalizing the solution in Lemma \ref{lemma:margin_gap} by its margin, denoted by $\alpha\geq 1-O(d^{-1/10})$,  the solution to \eqref{equ:local_SL_equi} (which should have no-larger norm) satisfies
\#\label{equ:local_SL_equi_norm}
\sum_{i=1}^{2K}\|\tilde w_{i}\|_2^2\leq \sum_{i=1}^{2K}\Big\|\frac{w_{i}^*}{\alpha}\Big\|_2^2=\frac{2+(2K-2)\cdot  O(d^{-3/10})}{\alpha^2}\le 2+  O(d^{-\frac{1}{10}}). 
\#
On the other hand, continue to follow the argument of Eq. (21)-(28) in the proof of Theorem 3.1 in \cite{liu2021self}, we know that for any $\ell\in[2K]\setminus\{1,2\}$, 
\#\label{equ:one_j_bound}
\la\tilde w_1,e_1 \rangle^2+\la\tilde w_2,e_1 \rangle^2 + \la\tilde w_\ell,e_1 \rangle^2 \geq 2-O(d^{-1/10}).
\#
Note that this also implies 
\#\label{equ:all_j_bound}
\sum_{i=1}^{2K}\|\tilde w_i\|_2^2\geq 2-O(d^{-1/10}).
\#
By \eqref{equ:local_SL_equi_norm} and \eqref{equ:one_j_bound}, we know that
\$
\sum_{j=2}^d\Bigg(\la\tilde w_1,e_j \rangle^2+\la\tilde w_2,e_j \rangle^2 + \la\tilde w_\ell,e_j \rangle^2\Bigg) \leq O(d^{-1/10})
\$
which further leads to the fact that 
\#\label{equ:one_j_bound_1}
&\sum_{j=2}^d\Bigg(\la\tilde w_1,e_j \rangle^2+\la\tilde w_2,e_j \rangle^2 + \sum_{\ell\in[2K]\setminus\{1,2\}}\la\tilde w_\ell,e_j \rangle^2\Bigg)\\
&\leq \sum_{\ell\in[2K]\setminus\{1,2\}}\sum_{j=2}^d\Bigg(\la\tilde w_1,e_j \rangle^2+\la\tilde w_2,e_j \rangle^2 + \la\tilde w_\ell,e_j \rangle^2\Bigg) \leq 2K\cdot O(d^{-1/10})\leq O(d^{-1/20}).\notag
\#
The rest of the proof follows that of Theorem 3.1 in \cite{liu2021self}, with the number of classes $3$ therein being replaced by $c=2K$ (as Lemma E.3 in \cite{liu2021self} still holds). By applying the argument therein for all $e_j$ with $j=2,\cdots,d$, we have 
\#\label{equ:ej_corr_final_full}
\sum_{j=2}^d \sum_{i=1}^m\langle u_{i}, e_{j} \rangle^2\le \sum_{j=2}^d \sum_{\ell\in[2K]}\la\tilde w_\ell,e_j \rangle^2\leq  O(d^{-\frac{1}{20}}). 
\# 

Furthermore, notice that by Lemma E.3 in \cite{liu2021self}, $u(u)^\top =(v)^\top v$ at the solution and $\|\tilde w\|_F^2=2\cdot  \|u(u)^\top\|_F^2$. 
Hence
\$
\bigg(\sum_{i=1}^m \|u_i\|_2^2\bigg)^2= \|u\|_F^4\geq \|u(u)^\top\|_F^2= \|\tilde w\|_F^2/2\geq 1-O(d^{-1/10}),
\$
where the last inequality uses \eqref{equ:all_j_bound}. 
This leads to the final result that
\$
\sum_{i=1}^m\langle u_{i}, e_{1} \rangle^2\ge \sum_{i=1}^m \|u_i\|_2^2 - O(d^{-\frac{1}{20}})\geq 1-O(d^{-1/20}),
\$ 
where we use $K= \Theta(d^{1/20})$ and \eqref{equ:ej_corr_final_full}. Note that the proof above also holds for other dataset $k\neq 1$.  This completes the  proof. 
\hfill\qed 


 

\end{document}